\documentclass[letterpaper]{article} 
\usepackage{aaai2026}  
\usepackage{times}  
\usepackage{helvet}  
\usepackage{courier}  
\usepackage[hyphens]{url}  
\usepackage{graphicx} 
\usepackage{natbib}  
\usepackage{caption} 
\usepackage{algorithm}
\usepackage{algorithmic}

\usepackage{amsmath}
\usepackage{amssymb}
\usepackage{booktabs} 
\usepackage{multirow}
\usepackage[table]{xcolor} 
\usepackage{pifont} 
\usepackage{graphicx}  
\usepackage{subcaption} 

\newcommand{\ResetAppendixCounters}{%
\setcounter{equation}{0}
\setcounter{figure}{0}
\setcounter{table}{0}
}

\newcommand{\AppendixIndependentNumbering}{%
\ResetAppendixCounters 
\renewcommand{\theequation}{A.\arabic{equation}}
\renewcommand{\thefigure}{A.\arabic{figure}}
\renewcommand{\thetable}{A.\arabic{table}}
\renewcommand{\thesection}{A.\arabic{section}}
\renewcommand{\thesubsection}{A.\arabic{section}.\arabic{subsection}}
}

\newcommand{\AppendixSection}[1]{%
  \refstepcounter{section}
  \section*{\thesection\ #1}
}

\newcommand{\AppendixSubsection}[1]{%
  \refstepcounter{subsection}
  \subsection*{\thesubsection\ #1}
}

\usepackage{newfloat}
\usepackage{listings}
\DeclareCaptionStyle{ruled}{labelfont=normalfont,labelsep=colon,strut=off} 
\floatstyle{ruled}
\newfloat{listing}{tb}{lst}{}
\floatname{listing}{Listing}
\title{Wasserstein-Aligned Hyperbolic Multi-View Clustering}
\author {
    Rui Wang\textsuperscript{\rm 1}, 
    Yuting Jiang\textsuperscript{\rm 1},
    Xiaoqing Luo\textsuperscript{\rm 1},
    Xiao-Jun Wu\textsuperscript{\rm 1},
    Nicu Sebe\textsuperscript{\rm 2},
    Ziheng Chen\textsuperscript{\rm 2}\thanks{Corresponding author.}
}
\affiliations {
    \textsuperscript{\rm 1}School of Artificial Intelligence and Computer Science, Jiangnan University \\
    \textsuperscript{\rm 2}Department of Information Engineering and Computer Science, University of Trento\\
    \fontsize{10pt}{12pt}\selectfont
    \{cs\_wr, xqluo, wu\_xiaojun\}@jiangnan.edu.cn, 
    \{yuting\_jiang2025, ziheng\_ch\}@163.com
}

\begin{document}
\maketitle
\begin{abstract}
Multi-view clustering (MVC) aims to uncover the latent structure of multi-view data by learning view-common and view-specific information. Although recent studies have explored hyperbolic representations for better tackling the representation gap between different views, they focus primarily on instance-level alignment and neglect global semantic consistency, rendering them vulnerable to view-specific information (\textit{e.g.}, noise and cross-view discrepancies). To this end, this paper proposes a novel Wasserstein-Aligned Hyperbolic (WAH) framework for multi-view clustering. Specifically, our method exploits a view-specific hyperbolic encoder for each view to embed features into the Lorentz manifold for hierarchical semantic modeling. Whereafter, a global semantic loss based on the hyperbolic sliced-Wasserstein distance is introduced to align manifold distributions across views. This is followed by soft cluster assignments to encourage cross-view semantic consistency. Extensive experiments on multiple benchmarking datasets show that our method can achieve SOTA clustering performance. 
\end{abstract}

\begin{links}
    \link{Code}{https://github.com/Yuting-jiang-jnu/WAH-MVC}
\end{links}

\section{Introduction}
Multi-view data consists of heterogeneous features or originates from multiple sources. Although describing the same underlying semantics, each view provides complementary information, capturing different aspects of the data. Integrating multiple views can significantly boost clustering performance~\cite{Chen2022ASR,xu2022self, 9738595}.
To this end, multi-view clustering (MVC) aims to exploit both the consistency and complementarity among views to achieve more accurate clustering. Traditional MVC approaches fall into several families, including subspace learning~\cite{Chen2021MLRR,tao2021ensemble, Xie2024SSCNet}, nonnegative matrix factorization~\cite{hu2019one,wei2022multiview}, graph-based methods~\cite{li2021incomplete,Chen2022ASR}, and kernel fusion~\cite{liu2018late,liu2021efficient}. 
While effective in low-dimensional scenarios, these shallow models struggle with scalability and complex data due to limited representational power.

To overcome these limitations, deep multi-view clustering (DMVC) methods have emerged as a promising paradigm that harnesses the representation power of deep neural networks to learn the latent embeddings from multiple views~\cite{li2019deep,li2022twin,wang2022adversarial,zhang2024learning,guo2024robust}. These methods aim to simultaneously capture view-specific semantics while jointly discovering a shared clustering structure in the latent space. \citet{xu2022self} introduce a self-supervised autoencoder with view consensus regularization, while \citet{zhang2024learning} impose hierarchical latent constraints to enhance semantic consistency. However, reconstruction-based models may struggle to learn discriminative representations. To address this issue, \citet{wang2022adversarial} employs adversarial strategies, combining view-specific encoders with discriminators to tackle this issue. However, adversarial training remains challenging due to its inherent instability and sensitivity, particularly in scenarios where the quality or completeness of views varies significantly \cite{xing2021algorithmic,xiao2022stability}.
\begin{figure}[t]
    \centering
    \includegraphics[width=1\linewidth]{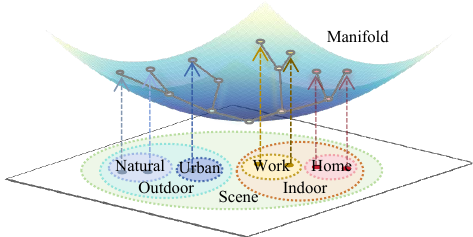}
    \caption{Real-world scene categories often form latent hierarchies. Here, fine-grained labels (\textit{e.g.}, Home, Work, Natural, Urban) are grouped into higher-level concepts like Indoor and Outdoor. Hyperbolic space has shown success in modeling hierarchical structure.}
    \label{fig: tree}
\end{figure}

In contrast, contrastive learning has emerged as a more direct and stable alternative to enforce cross-view alignment~\cite{peng2022xai, li2022twin, cui2024dual, zhang2025incomplete, Hu_Tian_Liu_Ye_2025}. It encourages semantically similar samples from different views to reside nearby in the latent space, while preserving the diversity of view-specific representations. Notably, \citet{peng2022xai} disentangles view-invariant and view-specific components via contrastive objectives, whereas \citet{cui2024dual} develops a dual-level contrastive framework that jointly aligns instance-level and cluster-level semantics. However, these methods are typically confined to Euclidean geometry, which limits their ability to capture non-Euclidean structures such as the hierarchical structure inherent in many real-world datasets (as illustrated in Figure~\ref{fig: tree}).

\begin{figure}[t]
    \centering
    \includegraphics[width=1\linewidth]{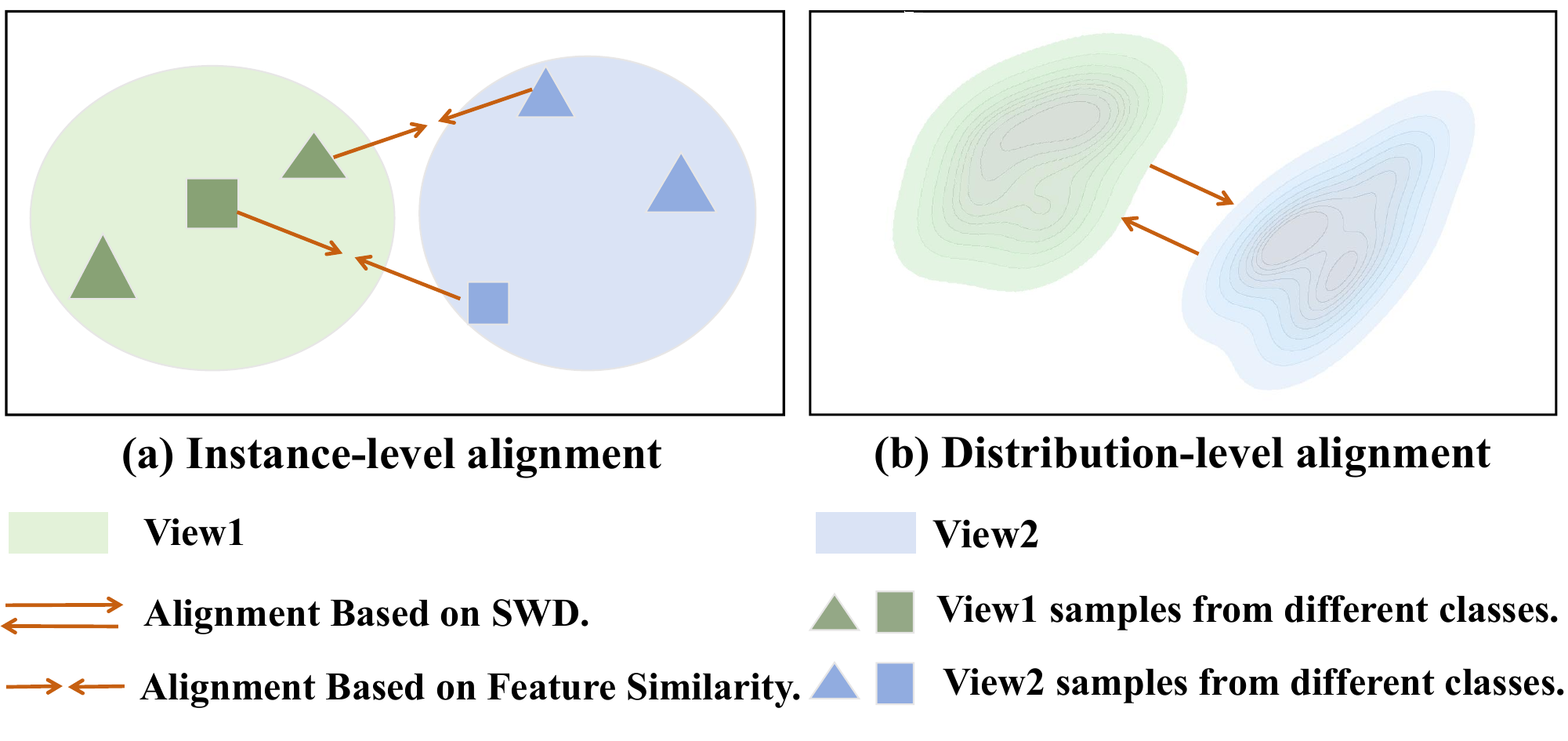}
    \caption{
    Comparison of traditional instance-level alignment and the proposed hyperbolic distribution-level alignment. Instance-level alignment maximizes pairwise similarity but poorly captures global semantics, whereas our method aligns holistic view-wise distributions via sliced-Wasserstein distance (SWD) on the Lorentz manifold.
    }
    \label{fig:sample_distribution}
\end{figure}

In response, recent efforts have extended contrastive learning to non-Euclidean spaces to better capture the underlying geometric structures of complex data. 
\citet{lin2022contrastive, Lin2023MHCN}, for instance, embed features onto hyperbolic manifolds to model latent semantics with greater geometric fidelity. 
However, existing methods align views at the instance level, aiming to pull corresponding samples from different views closer while pushing apart unrelated ones based on feature similarity, as shown in Figure~\ref{fig:sample_distribution}a. Although such a strategy is effective in aligning individual data points, it often overlooks the global semantic consistency across views. Moreover, it relies on manually selected positive and negative pairs, which introduces sampling bias and limits scalability when dealing with data containing a wide range of variations. To address these limitations, Optimal Transport (OT) has been introduced as a powerful tool for aligning global feature distributions under different views~\cite{zhang2024learning}. 
Unlike pointwise methods, distribution-level alignment via OT, most commonly quantified by the Wasserstein distance, captures holistic semantic correspondences by matching entire sample distributions. This approach effectively mitigates view gaps and enhances the learning of cross-view common semantics.
However, generalizing Wasserstein distance to non-Euclidean geometries remains a challenging problem, as it requires preserving the intrinsic geometric structure of the data while maintaining computational efficiency.

In this paper, we propose WAH-MVC, a unified Wasserstein-Aligned Hyperbolic framework for multi-view clustering that simultaneously preserves hierarchical structures, aligns global manifold distributions, and enhances feature discriminability. 
Unlike prior methods based on the Poincaré model, we adopt the Lorentz model for its closed-form Riemannian operations and superior numerical robustness, which facilitate stable optimization and seamless integration with deep learning frameworks~\cite{nickel2018learning,chami2020low}.
Specifically, WAH-MVC leverages multiple hyperbolic autoencoders to extract view-specific hierarchies and introduces a hyperbolic sliced-Wasserstein distance (SWD)-based constraint to achieve global distribution alignment across views in a geometry-aware manner, as illustrated in Figure~\ref{fig:sample_distribution}b.
On top of the learned embeddings, soft semantic labels are inferred via Lorentz Multinomial Logistic Regression (MLR), which maintains geometric consistency with the hyperbolic space and yields more suitable decision boundaries for curved latent structures. These labels further guide contrastive and consistency constraints, enabling the model to capture shared semantic information across multiple views.

Our main contributions are summarized as follows:
\begin{itemize}
\item \textbf{A new hyperbolic MVC framework:} We propose WAH-MVC to jointly preserve hierarchical structures, align global manifold distributions, and enhance feature discriminability on the Lorentz manifold of hyperbolic space.

\item \textbf{A novel Lorentz alignment mechanism:} We introduce an Alignment mechanism based on the Lorentz SWD, enabling effective distribution-level alignment across views.

\item \textbf{Experimental effectiveness:} Extensive empirical evaluations show the superiority of WAH-MVC over several SOTA methods in MVC.
\end{itemize}

\section{Preliminary}
In this section, we give a brief introduction to the Lorentz model of hyperbolic space and the Wasserstein Distance on Riemannian manifolds. 
\subsection{Hyperbolic Lorentz Model}
The $n$-dimensional Lorentz model of hyperbolic space with constant negative curvature $K<0$ is defined as
\begin{equation}
    \mathbb{L}_K^n = \left\{ x \in \mathbb{R}^{n+1} \mid \langle x, x \rangle_{\mathcal{L}} = \frac{1}{K}, \, x_0 > 0 \right\},
\end{equation}
where $\langle x, y \rangle_{\mathcal{L}} = -x_0 y_0 + \sum_{i=1}^n x_i y_i$ denotes the Lorentz inner product. The geodesic distance between two points is given by
\begin{equation}
d_{\mathbb{L}}(x, y) = \frac{1}{\sqrt{|K|}} \cdot \operatorname{arccosh}\left( |K| \langle x, y \rangle_{\mathcal{L}} \right),
\end{equation}
where $x, y \in \mathbb{L}_K^n$. In our work, we adopt the canonical origin $\mathbf{x}_\mathrm{o} = [\sqrt{-1/K},\, 0,\, \ldots,\, 0] $ as the reference point for tangent-space operations. At $\mathbf{x}_\mathrm{o}$, the tangent space $T_{\mathbf{x}_\mathrm{o}} \mathbb{L}_K^n$ is defined as the set of all vectors in $\mathbb{R}^{n+1}$ orthogonal to $\mathbf{x}_\mathrm{o}$ with respect to the Lorentz inner product.
The exponential map at $\mathbf{x}_\mathrm{o}$, which projects a tangent vector $v \in T_{\mathbf{x}_\mathrm{o}} \mathbb{L}_K^n$ onto the manifold, is defined as
\begin{equation}
    \exp_{\mathbf{x}_\mathrm{o}}^K(v) = \cosh(\alpha)\,\mathbf{x}_\mathrm{o} + \sinh(\alpha)\,\frac{v}{\alpha},
\end{equation}
where $\alpha = \sqrt{|K|}\,\|v\|_{\mathcal{L}}$ and $\|v\|_{\mathcal{L}} = \sqrt{\langle v, v \rangle_{\mathcal{L}}}$ denotes the Lorentz norm. We present only the Lorentz operators utilized in this paper, while other operators and their detailed formulations are provided in the Appendix~\ref{appendix: Lorentz}.


\begin{figure*}[t]
    \centering
    \includegraphics[width=\textwidth]{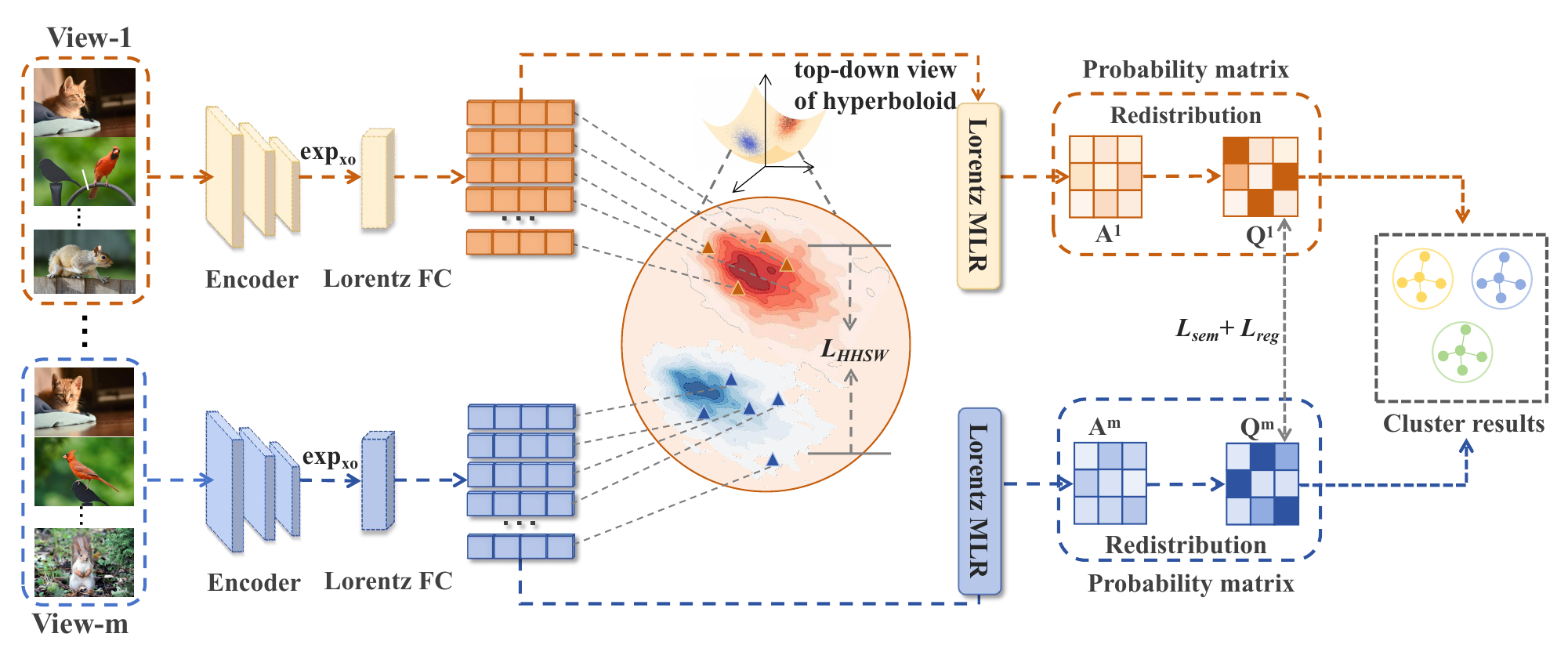}
    \caption{The framework of WAH-MVC. It consists of three main components: (1)
\textbf{Lorentz Feature Embedding}, which maps view-specific features into a curvature-aware Lorentz manifold for improved representation learning; (2) \textbf{Wasserstein-Alignment for Feature Distribution} ($\mathcal{L}_{\text{HHSW}}$), which aligns distributions of different views in hyperbolic space using a sliced Wasserstein distance; (3) \textbf{Contrastive Cluster Enhancement} ($\mathcal{L}_{\text{sem}} + \mathcal{L}_{\text{reg}}$), which enhances cluster discriminability by enforcing semantic consistency across views through contrastive learning.}

    \label{fig: framework}
\end{figure*}

\subsection{Wasserstein Distance on Riemannian Manifolds} 
Wasserstein distance serves as the core cost metric in OT, and has been successfully applied in various tasks involving distribution alignment and semantic matching. Let $\mu, \nu \in \mathcal{P}_p(\mathcal{M})$ be two probability measures defined on a Riemannian manifold $\mathcal{M}$. The $p$-Wasserstein distance between $\mu$ and $\nu$ is given by
\begin{equation}
\mathcal{W}_p(\mu, \nu) = \min_{\gamma \in \Pi(\mu, \nu)} \int_{\mathcal{M} \times \mathcal{M}} d_{\mathcal{M}}(\mathbf{x}, \mathbf{y})^p \, d\gamma(\mathbf{x}, \mathbf{y}),
\label{eq:ot}
\end{equation}
where $d_{\mathcal{M}}$ denotes the geodesic distance on $\mathcal{M}$, and $\Pi(\mu, \nu)$ is the set of all couplings (\textit{i.e.}, joint distributions) with marginals $\mu$ and $\nu$.
In our setting, we treat hyperbolic embeddings $\mathcal{Z}_{\text{hyp}} = \{\mathbf{z}_i\}_{i=1}^N$ on the manifold as an empirical measure $\mu = \frac{1}{N} \sum_{i=1}^N \delta_{\mathbf{z}_i}$, where $\delta_{\mathbf{z}}$ denotes the Dirac measure centered at $\mathbf{z}$. Based on this construction, feature alignment across views reduces to computing the Wasserstein distance between empirical distributions on the manifold.
\section{Method}
As shown in Figure~\ref{fig: framework}, the proposed WAH-MVC framework comprises three main components: Lorentz feature embedding, Wasserstein-Alignment for feature distribution, and contrastive clustering enhancement. The following subsections provide a detailed description.

\subsection{Lorentz Feature Embedding}
Given an $M$-view dataset $\mathcal{X} = \{X_1, \dots, X_M\}$, where $X_m = \{\mathbf{x}_i^m \in \mathbb{R}^{D_m}\}_{i=1}^N$, $N$ is the number of samples and $D_m$ the feature dimension of view $m$.
Our goal is to learn representations capturing both view-specific and shared information for clustering semantically consistent samples.

For each view $m$, we employ a geometry-aware encoder that maps the input features onto the Lorentz manifold. Specifically, a Euclidean encoder extracts latent features from the input $X_m$, which are then mapped to the $(d+1)$-dimensional Lorentz manifold via the exponential map $\exp_{\mathbf{x}_\mathrm{o}}(\cdot)$ at the Lorentz origin $\mathbf{x}_\mathrm{o}$. To further reduce dimensionality and enhance discriminability,we apply a Lorentz Fully Connected (FC) layer with curvature-aware normalization ~\cite{bdeir2023fully, chen2025gyrobn, chen2025riemannian} within the Lorentz space. The overall transformation can be written as:
\begin{equation}
 \tilde{\mathcal{Z}}_{\text{hyp}}^{(m)} = \text{LorentzFC} \left( \exp_{\mathbf{x}_\mathrm{o}}(f_{\mathrm{enc}}(X_m; \theta_{\mathrm{enc}}^m)) \right),
\end{equation}
where $f_{\mathrm{enc}}(\cdot)$ is the Euclidean encoder with view-specific parameters $\theta_{\mathrm{enc}}^m$.
The Lorentz FC is a generalization of the Euclidean FC layer to Lorentzian geometry (detailed in App.~\ref{appendix: Lorentz}).
This process yields the final hyperbolic embeddings, denoted by $\tilde{\mathcal{Z}}_{\text{hyp}}^{(m)} = \{\tilde{\mathbf{z}}_i^{m} \in \mathbb{L}_K^{r} \}_{i=1}^N$.

\subsection{Wasserstein-Alignment for Feature Distribution}
To enhance semantic consistency across views in hyperbolic space, we propose a global Wasserstein-Alignment strategy that operates directly on the Lorentz manifold. However, computing Wasserstein distance using Eq.~\ref{eq:ot} in curved spaces such as hyperbolic spaces poses significant computational challenges due to the nonlinearity of geodesic distances and the complexity of coupling space. 
As a countermeasure, we adopt a scalable approximation based on SWD. Specifically, following Bonet et al.~\cite{bonet2023hyperbolic}, we leverage two hyperbolic extensions of Horospherical Hyperbolic Sliced-Wasserstein (HHSW) and Geodesic Hyperbolic Sliced-Wasserstein (GHSW)—both of which are formulated within the Lorentz model. In this work, we focus on the HHSW variant due to its favorable computational properties and compatibility with our framework.

Formally, for probability measures $\mu, \nu \in \mathcal{P}_{p}(\mathbb{L}_K^n)$ on the Lorentz model, the HHSW distance is defined as
{\small
\begin{equation}
\text{HHSW}_{\theta}^p(\mu, \nu) = \int_{T_{\mathbf{x}_\mathrm{o}}\mathbb{L}_K^n \cap \mathbb{S}^{n-1}} \mathcal{W}_p(B_{\theta}\#\mu, B_{\theta}\#\nu) \, d\lambda(\theta),
\end{equation}
}
where $\mathbf{x}_\mathrm{o}$ is the lorentz origin, and $\mathbb{S}^{n-1}$ is the unit sphere in the tangent space $T_{\mathbf{x}_\mathrm{o}}\mathbb{L}_K^n$. The operator $B_{\theta}$ denotes the Busemann projection along the horospherical direction $\theta$, which maps points in $\mathbb{L}_K^n$ to scalar values in $R$. The symbol $\mathcal{W}_p$ represents the standard $p$-Wasserstein distance in the real line. Since the projected distributions $B_{\theta}\#\mu$ and $B_{\theta}\#\nu$ are one-dimensional, $\mathcal{W}_p$ can be computed efficiently using inverse cumulative distribution functions:
{\small
\begin{equation}
\mathcal{W}_p(B_{\theta}\#\mu, B_{\theta}\#\nu) = \int_0^1 \left| F^{-1}_{B_{\theta}\#\mu}(u) - F^{-1}_{B_{\theta}\#\nu}(u) \right|^p du ,
\label{eq:W_p}
\end{equation}
}
with $F^{-1}$ denoting the quantile function. The integration over directions $\theta$ is taken for the uniform measure $\lambda$ on the unit sphere.
This sliced formulation preserves the curvature of the hyperbolic space while greatly reducing computational complexity. 
As a result, it enables efficient and scalable alignment of global distributions across views, thus being well-suited for high-dimensional MVC tasks.

To implement HHSW in our multi-view setting, we first sample projection directions from $T_{\mathbf{x}_\mathrm{o}}\mathbb{L}_K^n$. Given a batch of features from the $m$-th view $\tilde{\mathcal{Z}}_{hyp}^{(m)}$, we generate a set of $L$ directions $\boldsymbol{\Theta}={\{\boldsymbol{\theta}_{\ell}}\}_{\ell=1}^L$, where $\theta_{\ell}$ satisfies the Lorentz orthogonality constraint, \textit{i.e.}, 
$
\langle \theta_\ell, \theta_\ell \rangle_{\mathbb{L}} = 1, \langle \theta_\ell,\mathbf{x}_\mathrm{o }\rangle_{\mathbb{L}} = 0
$.
This ensures that the directions lie on the unit sphere and respect the geometry of the data manifold.

Then, each feature point $\tilde{z}_i^{m} \in \tilde{\mathcal{Z}}_{hyp}^{(m)}$ is projected onto the real line via a Buseman function, yielding a scalar that reflects its position along the geodesic defined by direction $\theta_\ell$. The resulting set forms a one-dimensional distribution $B_{\theta_\ell}^m$ for each view. The Buseman function is detailed in App.~\ref{appendix: HHSW}.

Whereafter, we compute the HHSW distance between every pair of views $(m, n) \in \mathcal{V}$ across all sampled directions to enforce multi-view alignment:
\begin{equation}
\text{HHSW}_{\theta_{\ell}}^p\left(\tilde{\mathcal{Z}}_{hyp}^{(m)}, \tilde{\mathcal{Z}}_{hyp}^{(n)}\right) = \mathcal{W}_p\left( B_{\theta_{\ell}}^{m}, B_{\theta_{\ell}}^{n} \right).
\end{equation}

Finally, the HHSW-based alignment loss is defined as the average discrepancy across all view pairs and projection directions, as illustrated below:
{
\small
\begin{equation}
\mathcal{L}_{\mathrm{HHSW}} = \frac{1}{|\mathcal{V}| \times L} \sum_{(m,n) \in \mathcal{V}} \sum_{\ell=1}^{L} \mathrm{HHSW}_{\theta_\ell}^p\left(\tilde{\mathcal{Z}}_{hyp}^{(m)} , \tilde{\mathcal{Z}}_{hyp}^{(n)} \right),
\end{equation}
}
where $\mathcal{V}$ is the set of unordered view pairs. This loss function provides a geometry-aware and computationally efficient mechanism for aligning multi-view representations on the Lorentz manifold.

\subsection{Contrastive Cluster Enhancement}
To improve clustering discriminability, we introduce a contrastive learning strategy that exploits cross-view semantic consistency. After aligning the Lorentz representations via the SWD-based method, we employ Lorentz MLR~\cite{bdeir2023fully} on the Lorentz embeddings to generate class probabilities. As a Lorentz extension of the Euclidean MLR, Lorentz MLR performs classification by defining distance-based decision boundaries that respect the curvature of the Lorentz manifold. The cluster probability matrix is computed by 
\begin{equation}
\mathcal{A}^{(m)} = \text{LorentzMLR}(\tilde{\mathcal{Z}}_{\text{hyp}}^{(m)}) = \{\mathbf{a}_i^m \in \mathbb{R}^K\}_{i=1}^N,
\end{equation}
where $K$ represents the number of clusters. The detailed formulation of LorentzMLR is provided in App.~\ref{appendix: Lorentz}.

To refine cluster assignments, we compute a target distribution $\mathbf{Q}^{(m)} \in \mathbb{R}^{N \times K}$ from $\mathcal{A}^{(m)}$, following the method proposed in~\cite{cui2023deep}:
{
\small
\begin{equation}
q_{ij}^{(m)} = \frac{(a_{ij}^{(m)})^2 / \sum_{i=1}^{N} a_{ij}^{(m)}}{\sum_{k=1}^{K} \left[(a_{ik}^{(m)})^2 / \sum_{i=1}^{N} a_{ik}^{(m)}\right]},
\label{eq:redistribution}
\end{equation}
}
Let $\mathbf{q}_{k}^{(m)}$ denote the $k$-th column of the matrix $\mathbf{Q}^{(m)}$, where $\mathbf{q}_{k}^{(m)} = [ q_{1k}^{(m)}, q_{2k}^{(m)}, \dots, q_{Nk}^{(m)} ]^\top$. Here, $q_{ij}^{(m)}$ denotes the probability of assigning sample $i$ to cluster $j$ in view $m$. 
This weighting mechanism boosts confident predictions while downplays ambiguous ones, thereby enabling more discriminative clustering.

Subsequently, the similarity between cluster-wise soft assignments is measured by the inner product, shown below:
\begin{equation}
s_{k,k}^{(m, n)} = \left(\mathbf{q}_{k}^{(m)}\right)^\top \mathbf{q}_{k}^{(n)}.
\end{equation}
Based on these cross-view similarities, a contrastive loss that encourages alignment between matched clusters while distinguishing mismatched ones is constructed.
Inspired by~\cite{chen2023deep}, the contrastive loss between views $m$ and $n$ is formulated as:
{
\small
\begin{equation}
\mathcal{L}_{\text{c}} = -\frac{1}{K} \sum_{k=1}^{K}
\log \frac{
\exp\left( s_{k,k}^{(m, n)} / \tau \right)
}{
\exp\left( s_{k,k} ^{(m,n)}/ \tau \right) +
\sum\limits_{\substack{j=1, j \ne k}}^{K} \exp\left( s_{k,j}^{(m,n)} / \tau \right)
},
\label{eq:label_contrastive_loss}
\end{equation}
}
with $\tau$ denoting a temperature parameter controlling the sharpness of the similarity distribution. Thereby, the total contrastive loss across all views is expressed as:
\begin{equation}
\mathcal{L}_{\text{sem}} = \sum_{m=1}^M \sum_{n=1,\, n \ne m}^M \mathcal{L}_{\text{c}}(m, n).
\end{equation}
To avoid degenerate assignments where all instances collapse into a single cluster, we follow \cite{cui2023deep} and incorporate a cross-view regularization term as follows:
\begin{equation}
\mathcal{L}_{\text{reg}} = \sum_{m=1}^{M} \sum_{j=1}^{K} p_j^{(m)} \log p_j^{(m)},
\end{equation}
where $p_j^{(m)}=\dfrac{1}{N} \sum_{i=1}^{N} q_{ij}^{(m)}$ is an entropy-based regularizer that encourages balanced cluster assignments across views, thus enhancing cross-view consistency.

\subsection{Optimization and Label Inference}
Previously, we have detailed the global alignment strategy via SWD and the semantic-aware contrastive learning module that captures shared semantics across views. Here, we summarize the overall training objective of WAH-MVC, which integrates three complementary loss functions:
\begin{equation}
\mathcal{L} = \alpha \mathcal{L}_{\text{HHSW}} + \beta \mathcal{L}{\text{sem}} + \gamma \mathcal{L}{\text{reg}},
\label{eq:total_loss}
\end{equation}
where $\alpha$, $\beta$, and $\gamma$ are three hyperparameters used to balance the contribution of each component.

After training, the final semantic label for the $i$-th ($i \in {1, 2, \dots, N}$) instance is predicted by averaging its soft assignments across all $M$ views and selecting the cluster with the highest mean probability, as shown below:
\begin{equation}
y_i = \underset{j}{\arg\max} \left( \frac{1}{M} \sum_{m = 1}^M q_{ij}^{(m)}\right).
\label{eq:end_label}
\end{equation}
This simple yet effective voting mechanism ensures robust prediction by leveraging consensus across views.

\definecolor{gray1}{gray}{0.90}
\begin{table*}[t]
\centering
{\fontsize{9}{10}\selectfont
\begin{tabular}{l | cc | cc| cc | cc | cc| cc}
\toprule
\multirow{3}{*}{{Method}}
& \multicolumn{2}{c|}{MNIST-USPS} 
& \multicolumn{2}{c|}{COIL-10} 
& \multicolumn{2}{c|}{Scene-15}
& \multicolumn{2}{c|}{Amazon}
& \multicolumn{2}{c|}{Fashion}
& \multicolumn{2}{c}{YoutubeVideo} \\
&   
\multicolumn{2}{c|}{(V=2, N=5000)} 
& \multicolumn{2}{c|}{ (V=3, N=720)} &
\multicolumn{2}{c}{(V=3, N=4485)} &
\multicolumn{2}{c|}{ (V=4,N=4790)}  & 
\multicolumn{2}{c|}{(V=3, N=10000)}  & \multicolumn{2}{c}{(V=3, N=101499)} \\

\cmidrule(lr){2-3} \cmidrule(lr){4-5} \cmidrule(lr){6-7}
\cmidrule(lr){8-9} \cmidrule(lr){10-11} \cmidrule(lr){12-13}
& ACC  & NMI  
& ACC  & NMI  
& ACC  & NMI  
& ACC  & NMI  
& ACC  & NMI  
& ACC  & NMI  
\\
\midrule
DSIMVC 
(ICML’22)
& 99.34& 98.13 
& 99.68& 99.35 
& 28.27& 29.04 
& 64.80 & 62.46 
& 95.50 & 92.17 
& 15.57& 7.46 
\\

MFLVC 
(CVPR’22)
& 99.56 & 98.73 
& 92.50 & 92.76 
& 34.94& 29.10 
& 89.44& 91.19 
& 99.25& 98.11 
& 21.55 & 22.88 
\\
DSMVC 
(CVPR’22)
& 96.34 & 94.27 
& 96.39& 95.32 
& 43.48 & 41.11 
& 38.64& 28.29 
& 79.40 & 77.98 
& 12.47 & 10.35 
\\

CVCL 
(ICCV’23)
& 99.70 & 99.13 
& 99.43 & 99.04 
& 44.59 & 42.17 
& 83.10 & 67.49 
& 99.31 & 99.21 
& 22.61 & 22.46 
\\
CPSPAN 
(CVPR’23)
& 93.30& 87.61 
& 82.92& 89.80 
& 22.92 & 15.69 
& 59.90& 53.30 
& 72.22 & 75.67 
& 23.88 & 22.24 
\\

GCFAgg 
(CVPR’23)
& 99.56 & 98.71 
& 85.83 & 94.08 
&42.27& 42.56 
& 90.58& 86.23 
& 98.97& 97.38 
& 27.58 &\textbf{27.31} 
\\
MVCAN 
(CVPR’24)
& 99.24 & 98.01 
& 99.31 & 98.57 
&38.32 &39.48
& 83.15 & 85.28 
& 82.89 & 86.14 
& 25.01 & 24.43 
\\

DCMVC 
(TIP’24)
& 95.76 & 90.27 
& 85.83 & 95.15 
& 32.60 & 28.34 
& 68.46 & 62.43 
& 94.89 & 92.42 
& 27.26 & 27.29 
\\

SCM 
(IJCAI’24)
& 98.94 & 97.02 
& 99.86 & 99.68 
&40.83&38.89 
& 48.98 & 40.20 
& 98.00 & 95.80 
& 18.10 & 18.49 
\\

CSOT 
(TIP’24)
& 99.22 & 99.22 
& 99.58 & 99.24 
& 35.83& 29.82 
& 98.04 & 95.39 
& 99.12 & 97.81 
& 21.11 & 21.57 
\\
\midrule
\rowcolor{gray1}
WAH-MVC 
& \textbf{99.88} & \textbf{99.62} 
& \textbf{99.91} & \textbf{99.74} 
& \textbf{74.38} & \textbf{74.68} 
& \textbf{99.88} & \textbf{99.62} 
& \textbf{99.74} &\textbf{ 99.29} 
& \textbf{29.40} & 22.74 
\\
\bottomrule
\end{tabular}%
}
\caption{Comparison of clustering performance on six  multi-view datasets ($V$: Number of Views; $N$: Samples per View)}
\label{tab:compare_performance_all}
\end{table*}


\section{Experiments}
\subsection{Experimental Settings}
\textbf{Datasets.}
We evaluate WAH-MVC on six publicly available multi-view datasets. MNIST-USPS~\cite{Peng2019:COMIC:full} is a two-view dataset, with each view containing 5,000 handwritten digit images from different domains. COIL-10~\cite{xu2021multiVAE} consists of 720 grayscale images from 10 categories, with each object captured from three different poses. Fashion~\cite{xiao2017fashionmnist} includes 10,000 samples from 10 clothing categories, each observed from front, side, and back views. 
Scene-15~\cite{li2005bayesian} comprises 4,485 images from 15 classes, each represented by three types of handcrafted features. Amazon~\cite{saenko2010adapting} contains 4,790 color images from 10 categories under four views. 
Youtube Video~\cite{madani2013using} is a large-scale dataset with 101,499 samples from 31 classes, each represented by three feature vectors of dimensions 512, 647, and 838.

\textbf{Comparative Methods.} The following methods are selected for comparison: 
view-level contrastive methods, such as DSIMVC~\cite{tang2022deepa}, CPSPAN~\cite{jin2023deep}, MFLVC~\cite{Xu_2022_CVPR}, and DCMVC~\cite{cui2024dual}, aim to maximize feature consistency across views; clustering-level contrastive methods, including CVCL~\cite{chen2023deep} and SCM~\cite{luo2024simple}, focusing on enhancing clustering structure through assignment-level contrast; aggregation-based methods, such as DSMVC~\cite{tang2022deepb}, GCFAgg~\cite{Yan_2023_CVPR}, and MVCAN~\cite{xu2024investigating}, integrate multi-view features with adaptive or global strategies. We also include an OT-based method, CSOT~\cite{zhang2024learning}, improving MVC performance via Euclidean-based semantic alignment.

\textbf{Network Architecture and Parameter Settings.}
The proposed model employs a hybrid  Euclidean encoder $f_{\mathrm{enc}}(\cdot)$ to process both vector and image inputs. For the $m$-th vector view, features are encoded using a ReLU-activated MLP with 3–5 hidden layers. A typical 3-layer configuration is $[D_m, 256, 512, d]$, where $D_m$ denotes the input dimension and $d$ represents the shared output size. For the $m$-th image view, a lightweight CNN is employed, consisting of two convolutional blocks, an adaptive pooling layer,  and 2–3 fully connected layers. The final output is a feature vector with dimension $d$, selected from $\{256,\ 512,\ 1024\}$. The features are mapped onto the Lorentz manifold via $\exp_{\mathbf{x}_\mathrm{o}}(\cdot)$ and transformed by a LorentzFC layer to produce the final hyperbolic embeddings (see App.~\ref{appendix: Implementation Details} for details).
The total loss is weighted by hyperparameters $\alpha$, $\beta$, and $\gamma$, selected using grid search from the set $\{0.001, 0.005, 0.01, 0.05, 0.1, 1.0\}$. A temperature parameter $\tau$ controls the sharpness of the contrastive objective. Its selection process is detailed in the following Ablation Studies section.

\subsection{Performance Evaluation}

Table~\ref{tab:compare_performance_all} summarizes the clustering performance of different methods on six benchmark datasets, with the best results highlighted in bold. Overall, WAH-MVC consistently surpasses the baselines, demonstrating strong effectiveness in multi-view clustering.

On relatively simple and balanced datasets (MNIST-USPS, COIL-10, Fashion), the most recent methods perform competitively. Nonetheless, WAH-MVC consistently outperforms strong baselines (e.g., CVCL, SCM, CSOT) across all metrics. On more challenging datasets such as Scene-15 and Amazon, WAH-MVC achieves substantial improvements, surpassing the second-best methods by approximately 29.8\% (ACC) and 32.1\% (NMI) on Scene-15. Despite the inherent challenges of the YoutubeVideo dataset, including substantial inter-view heterogeneity, label noise, and large-scale complexity, WAH-MVC achieves the highest ACC among all compared methods. Specifically, it outperforms the previous SOTA method GCFAgg by 1.82\% in ACC, showcasing its good generalization ability. Although WAH-MVC reports lower NMI than GCFAgg, this may be attributed to NMI's sensitivity to fragmented or noisy labels~\cite{VinhNMI2010}. In contrast, the ACC metric captures overall assignment accuracy and may better reflect WAH-MVC’s ability to preserve dominant semantics under noisy ground truth.

These results collectively highlight two major advantages of WAH-MVC. Firstly, unlike contrastive learning methods such as CVCL and DSMVC that operate in Euclidean space with limited ability to model the intrinsic hierarchical structures of multi-view data, WAH-MVC leverages hyperbolic space to preserve hierarchy and effectively capture multi-level dependencies among views and clusters.
Secondly, the OT-based distribution alignment on the Lorentz manifold allows WAH-MVC to mitigate view discrepancies at the semantic level, thereby facilitating learning a more discriminative hyperbolic network embedding.

\definecolor{gray1}{gray}{0.90}
\begin{table}[t]
\centering
\setlength{\tabcolsep}{1mm}
{\fontsize{9}{10}\selectfont
\begin{tabular}{ccc|cc|cc|cc}
\toprule
\multirow{3}{*}{$\mathcal{L}_{\text{HHSW}}$} & 
\multirow{3}{*}{$\mathcal{L}_{\text{sem}}$} & 
\multirow{3}{*}{$\mathcal{L}_{\text{reg}}$} & 
\multicolumn{2}{c|}{Amazon} & 
\multicolumn{2}{c|}{Fashion} & 
\multicolumn{2}{c}{YoutubeVideo} \\
& & & 
\multicolumn{2}{c|}{\scriptsize (V=4, N=4790)} & 
\multicolumn{2}{c|}{\scriptsize (V=3, N=10000)} & 
\multicolumn{2}{c}{\scriptsize (V=3, N=101499)} \\
\cmidrule(lr){4-5} \cmidrule(lr){6-7} \cmidrule(lr){8-9} 
& & & ACC & NMI & ACC & NMI & ACC & NMI 
\\
\midrule
\textcolor{red}{\ding{55}} &\textcolor{green}{\ding{51}}  &\textcolor{green}{\ding{51}}  
& 99.38 & 98.37 & 99.55 & 98.81 & 25.49 & 21.46 \\
\textcolor{green}{\ding{51}}  & \textcolor{red}{\ding{55}} & \textcolor{green}{\ding{51}}  
& 27.57 & 20.23 & 27.28 & 15.53 & 13.26 & 3.41 \\
\textcolor{green}{\ding{51}}  &\textcolor{green}{\ding{51}}  & \textcolor{red}{\ding{55}} 
& 85.10 & 93.94 & 71.38 & 86.98 & 27.23 & 12.46 \\
\rowcolor{gray1}
\textcolor{green}{\ding{51}} &\textcolor{green}{\ding{51}} &\textcolor{green}{\ding{51}} 
& \textbf{99.88} & \textbf{99.62} & \textbf{99.74} & \textbf{99.29} & \textbf{29.40} & \textbf{22.74} \\
\bottomrule
\end{tabular}
}
\caption{Impact of each loss component.}
\label{tab:loss_ablation}
\end{table}

\definecolor{gray1}{gray}{0.90}
\begin{table}[t]
\centering
\setlength{\tabcolsep}{1mm}
{\fontsize{9}{10}\selectfont
\begin{tabular}{l | cc | cc | cc}
\toprule
\multirow{2}{*}{Method} 
& \multicolumn{2}{c|}{Amazon}
& \multicolumn{2}{c|}{Fashion}
& \multicolumn{2}{c}{YoutubeVideo} \\
\cmidrule(lr){2-3} \cmidrule(lr){4-5} \cmidrule(lr){6-7}
& ACC & NMI  
& ACC & NMI  
& ACC & NMI  
\\
\midrule
CVCL 
& 83.10 & 67.49 
& 99.31 & 98.21 
& 22.61 & 22.46 
\\
CSOT 
& 98.04 & 95.39 
& 99.12 & 97.81 
& 21.11 & 21.57 
\\
\rowcolor{gray1}
WAH-MVC w/ A
& 99.36 & 98.27 
& 99.23 & 98.04 
& 26.78 & 21.82 
\\
\rowcolor{gray1}
WAH-MVC w/ B
& 99.62 & 98.93 
& 99.22 & 98.01 
& 28.60& 22.20 
\\
\bottomrule
\end{tabular}%
}
\caption{Clustering performance with different WAH-MVC backbones.
A and B denote the encoders from CVCL and CSOT, respectively.}
\label{tab:backbone_performance}
\end{table}
\subsection{Ablation Studies}
In this part, we make a series of ablation studies to explore the impact of potential factors on the model performance. 

\textbf{Loss functions.}  
Here, we conduct clustering experiments on the Amazon, Fashion, and YoutubeVideo datasets to analyze the significance of each loss function in WAH-MVC. 
As shown in Table~\ref{tab:loss_ablation}, the full model that integrates all three losses consistently achieves the best performance across all datasets. Notably, removing $\mathcal{L}_{\text{sem}}$ results in a drastic drop in clustering quality, highlighting its necessity for semantic discriminability. The regularization term $\mathcal{L}_{\text{reg}}$ also enhances learning stability, especially on mid-scale datasets like Fashion. As the number of samples increases under a fixed number of views, the impact of $\mathcal{L}_{\text{HHSW}}$ becomes more pronounced. On the large-scale and highly heterogeneous YoutubeVideo dataset, incorporating $\mathcal{L}_{\text{HHSW}}$ respectively improves ACC and NMI by 3.91\% and 1.28\%, over the variant without using it. These findings demonstrate that $\mathcal{L}_{\text{HHSW}}$ effectively aligns multi-view distributions at the semantic level, thereby improving clustering performance.

\definecolor{gray1}{gray}{0.90}
\begin{table}[t]
\centering
{\fontsize{9}{10}\selectfont
\begin{tabular}{c | c| c c c}
\toprule
Dataset & Method & ACC & NMI & Time (s) \\
\midrule
\multirow{1}{*}{Scene-15}
& w/ $\mathcal{L}_{\text{HCL}} $& 72.27 & 73.32 & 5.95 \\
\multirow{1}{*}{(V=3, N=4485)}
&\cellcolor{gray1} w/ $\mathcal{L}_{\text{HHSW}} $ &\cellcolor{gray1}74.38 &\cellcolor{gray1}74.68 &\cellcolor{gray1}5.85 \\
\midrule
\multirow{1}{*}{Amazon}
& w/ $\mathcal{L}_{\text{HCL}}$  &99.53 &98.52&2.77 \\
\multirow{1}{*}{(V=4, N=4790)}
&\cellcolor{gray1} w/ $\mathcal{L}_{\text{HHSW}} $ &\cellcolor{gray1}99.78 &\cellcolor{gray1}99.39  &\cellcolor{gray1}2.67 \\
\midrule
\multirow{1}{*}{YoutubeVideo}
& w/ $\mathcal{L}_{\text{HCL}} $  & 22.45 & 23.19 & 53.32 \\
\multirow{1}{*}{(V=3, N=101499)}
&\cellcolor{gray1} w/ $\mathcal{L}_{\text{HHSW}} $ &\cellcolor{gray1}29.40 &\cellcolor{gray1}22.74 &\cellcolor{gray1}52.22 \\
\bottomrule
\end{tabular}
}
\caption{Clustering performance and training time comparison of $\mathcal{L}_{\text{HCL}}$ and $\mathcal{L}_{\text{HHSW}}$ loss-based hyperbolic alignment methods. Time: per-epoch training time (in seconds).}
\label{tab:CL_SWD_accuracy_time}
\end{table}

\begin{table}[t]
\centering
{\fontsize{9}{10}\selectfont
\begin{tabular}{l | cc | cc | cc}
\toprule
\multirow{2}{*}{Method}
& \multicolumn{2}{c|}{Fashion}
& \multicolumn{2}{c|}{Scene-15}
& \multicolumn{2}{c}{Amazon} \\
\cmidrule(lr){2-3} \cmidrule(lr){4-5} \cmidrule(lr){6-7}
& ACC& NMI 
& ACC & NMI 
& ACC &NMI 
\\
\midrule
GHSW
& 99.31 & 98.23 
& 72.77 & 73.47 
& 99.77 & 99.33 
\\
\midrule
\rowcolor{gray1} 
HHSW
& 99.74 & 98.29 
& 74.38 & 74.68 
& 99.88 & 99.62 
\\
\bottomrule
\end{tabular}
}
\caption{Projection Comparison: HHSW vs. GHSW}
\label{tab:hhsw_ghsw_performance}
\end{table}

\begin{figure}[t]
    \centering
    \includegraphics[width=\linewidth]{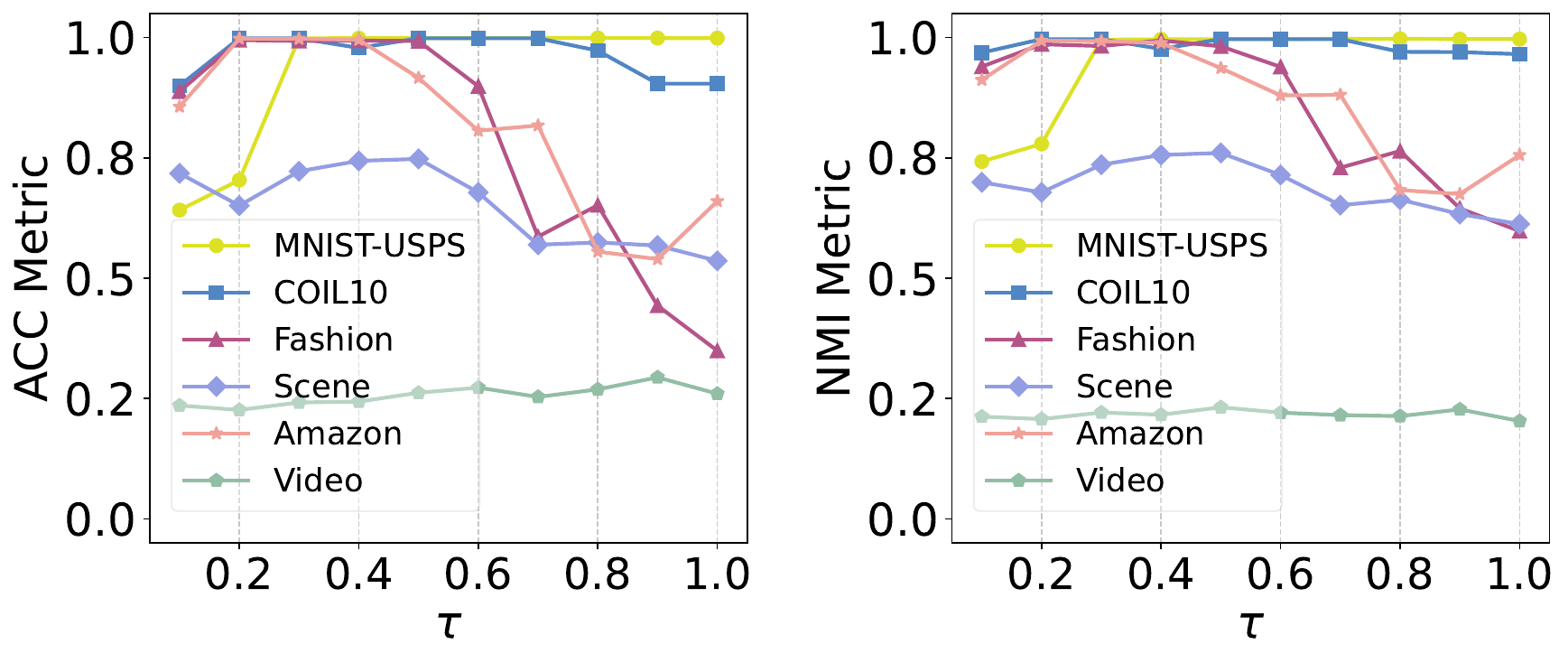}
   \caption{Effect of $\tau$ on ACC and NMI for all datasets.}

    \label{fig:temp_sensitivity_all}
\end{figure}

\textbf{Backbone models.}
To verify the robustness of WAH-MVC to different backbone architectures, we choose the encoders of CVCL and CSOT, which are used solely as feature extractors in our framework. 
As shown in Table~\ref{tab:backbone_performance}, WAH-MVC consistently achieves high clustering ACC and NMI on the Amazon and Fashion datasets, regardless of the encoder employed. This demonstrates that our model is insensitive to backbone variations. 
Even on the challenging YoutubeVideo dataset, where overall performance is lower, WAH-MVC still delivers notable improvements over the corresponding baseline models under the same encoders. 
Therefore, we argue that the strength of WAH-MVC stems from its core hyperbolic alignment and clustering mechanisms, rather than from a specific encoder design. 
\begin{figure}[t]
    \centering
    \includegraphics[width=\linewidth]{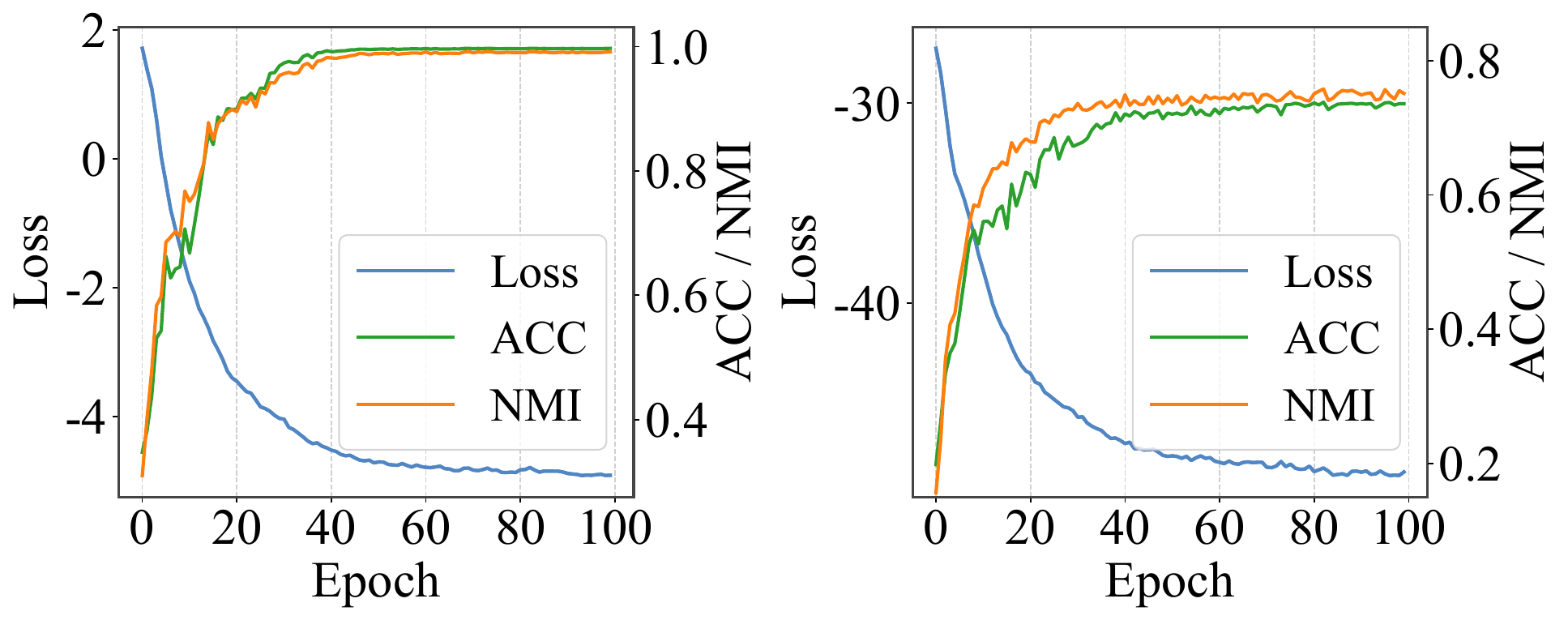}
    \caption{Convergence analysis of WAH-MVC on Amazon (left) and Scene-15 (right).}
    \label{fig:loss_analize}
\end{figure}


\begin{figure}[t]
    \centering
    \begin{subfigure}[t]{0.145\textwidth}
        \centering
        \includegraphics[width=\linewidth]{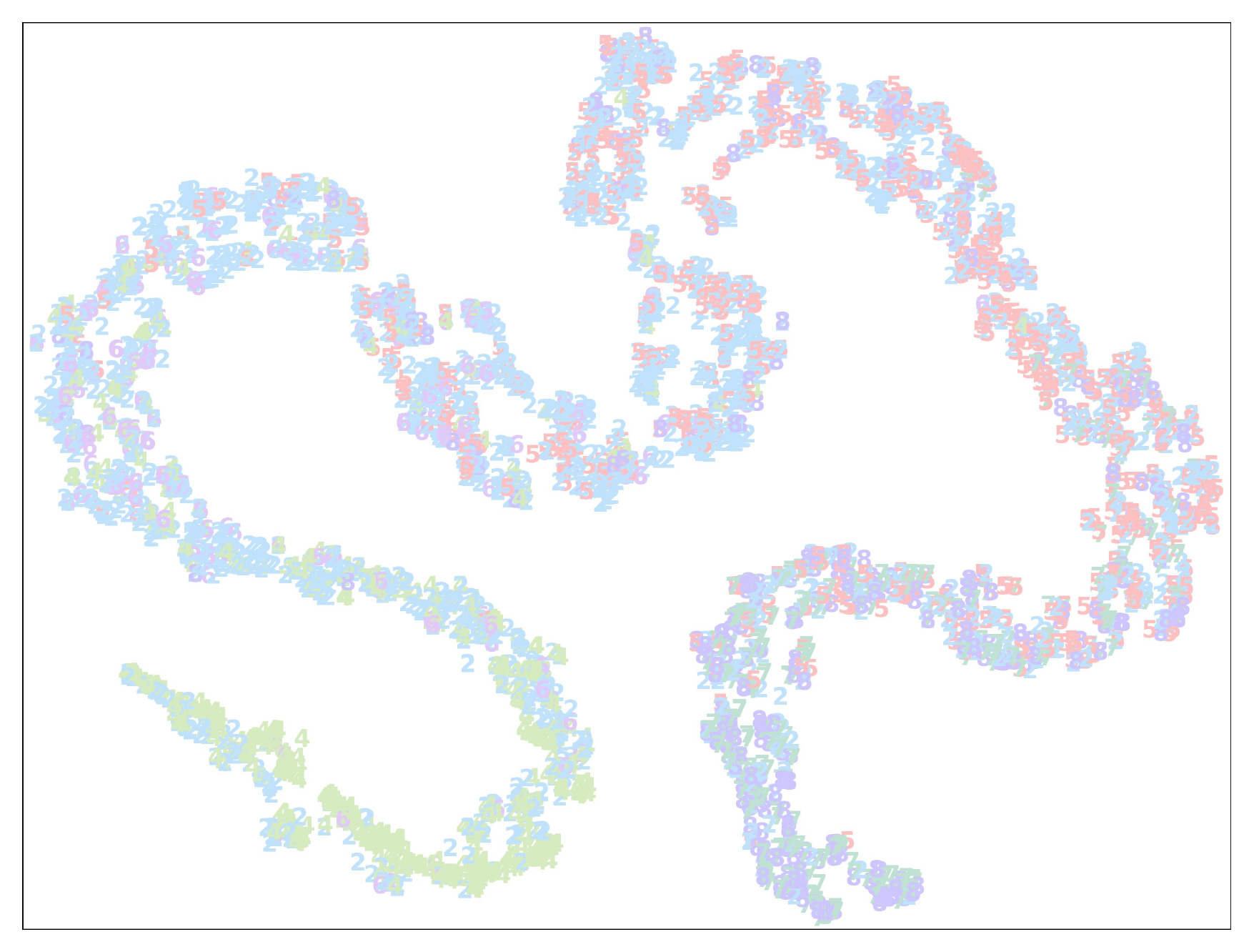}
        \caption{Epoch 0}
        \label{fig:Epoch0}
    \end{subfigure}
    \hspace{0.02mm}
    \begin{subfigure}[t]{0.145\textwidth}
        \centering
        \includegraphics[width=\linewidth]{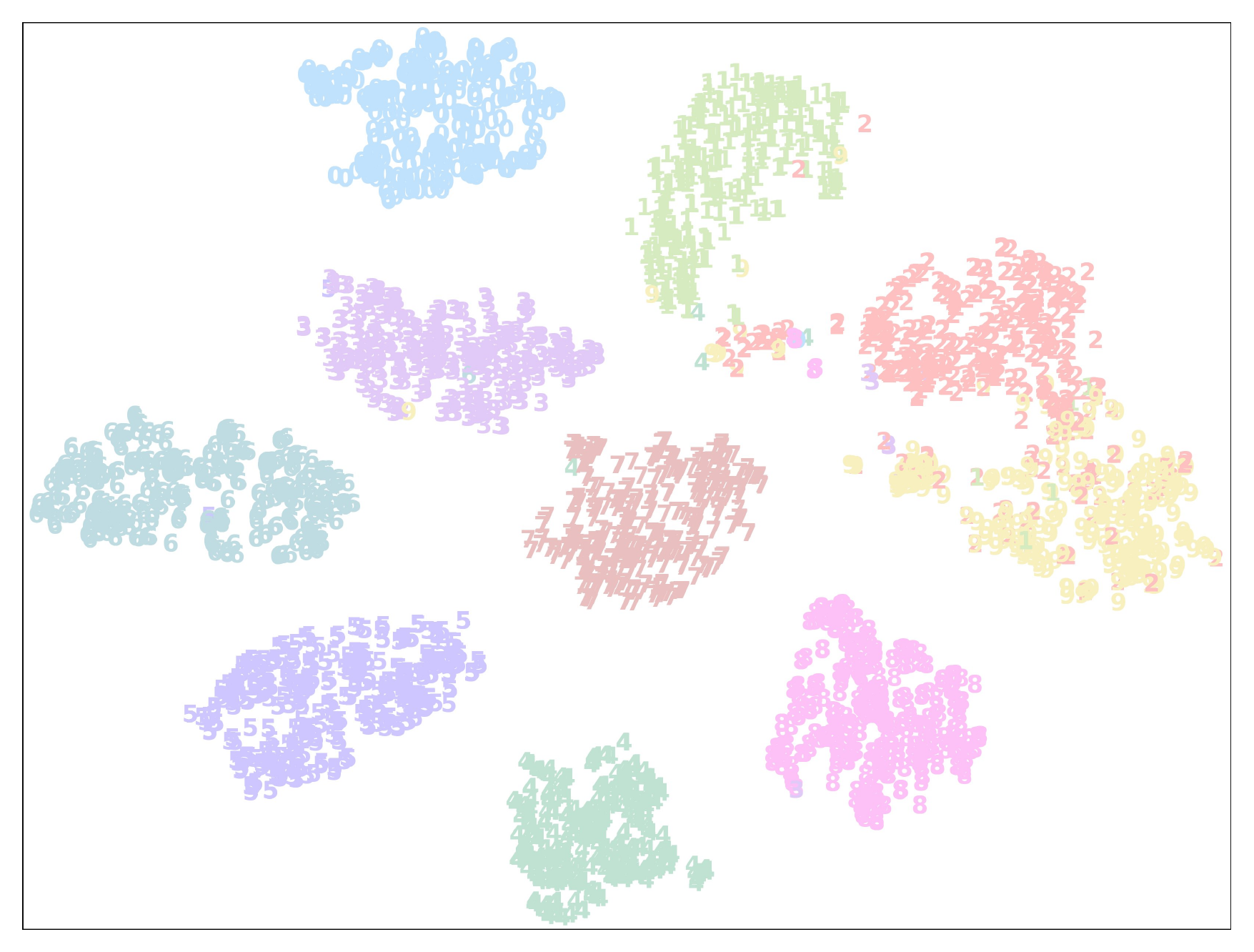}
        \caption{Epoch 40}
        \label{fig:Epoch40}
    \end{subfigure}
    \hspace{0.02mm}
    \begin{subfigure}[t]{0.145\textwidth}
        \centering
        \includegraphics[width=\linewidth]{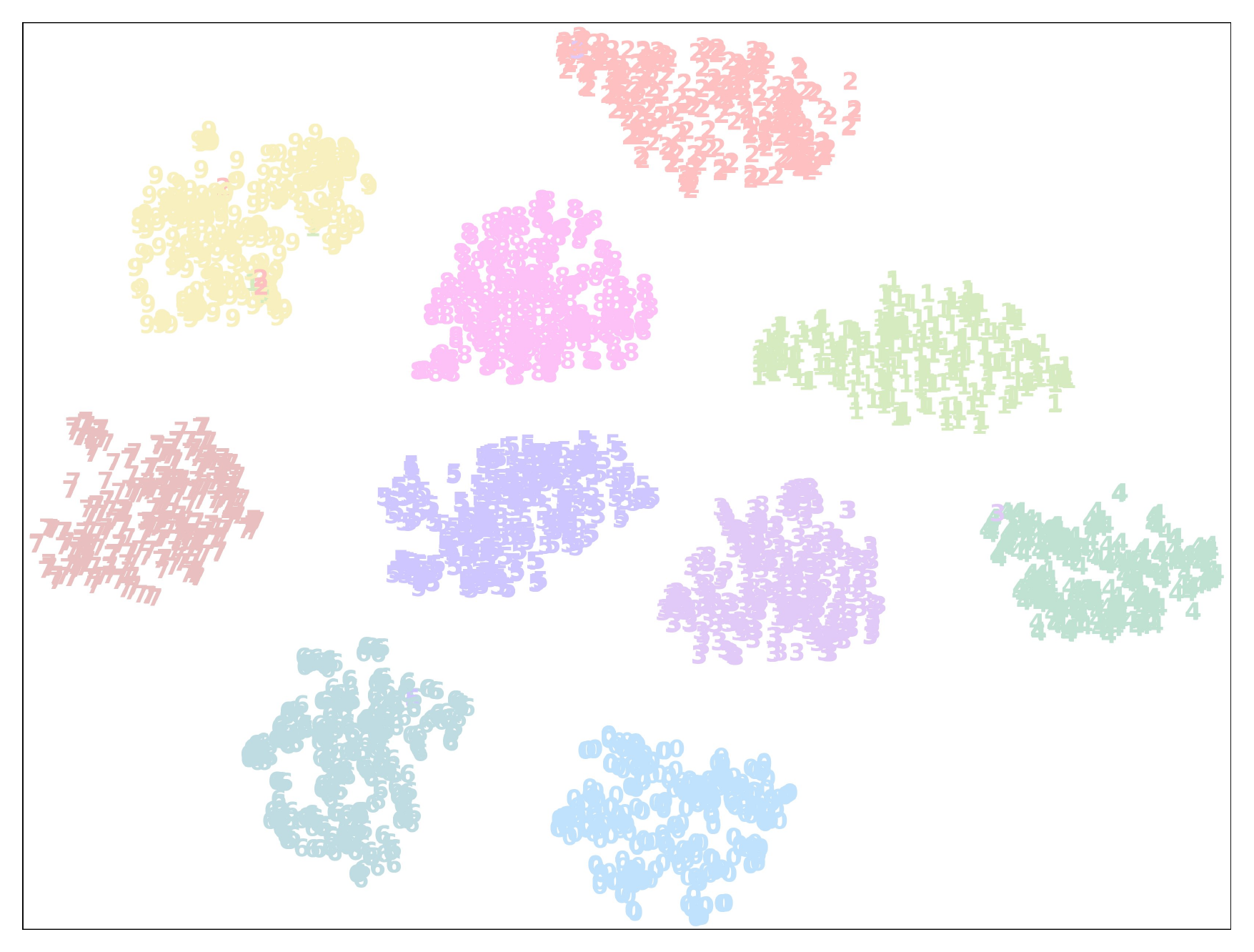}
        \caption{Epoch 80}
        \label{fig:Epoch80}
    \end{subfigure}

    \vspace{1mm} 

     \begin{subfigure}[t]{0.145\textwidth}
        \centering
        \includegraphics[width=\linewidth]{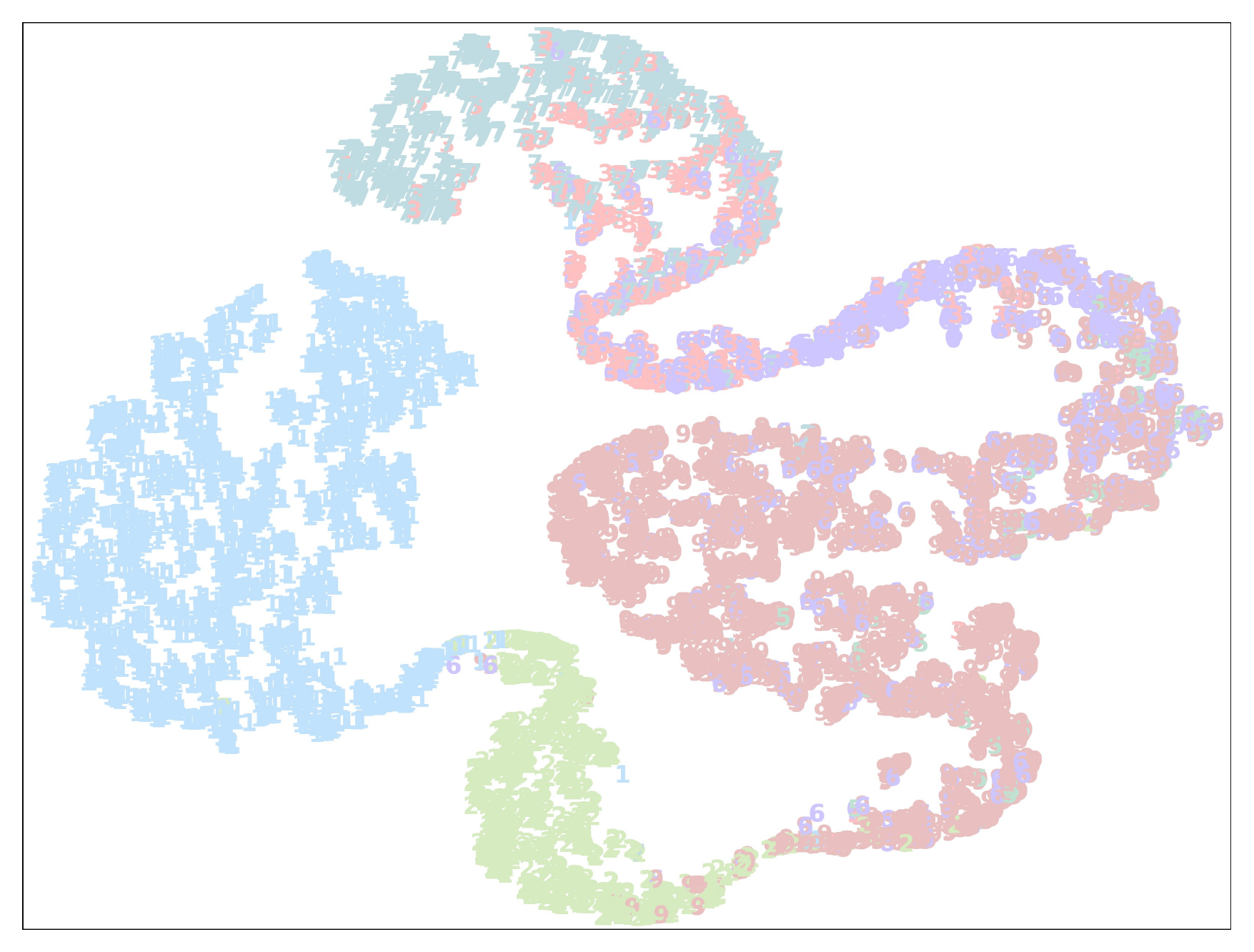}
        \caption{Epoch 0}
        \label{fig:Epoch01}
    \end{subfigure}
    \hspace{0.02mm}
    \begin{subfigure}[t]{0.145\textwidth}
        \centering
        \includegraphics[width=\linewidth]{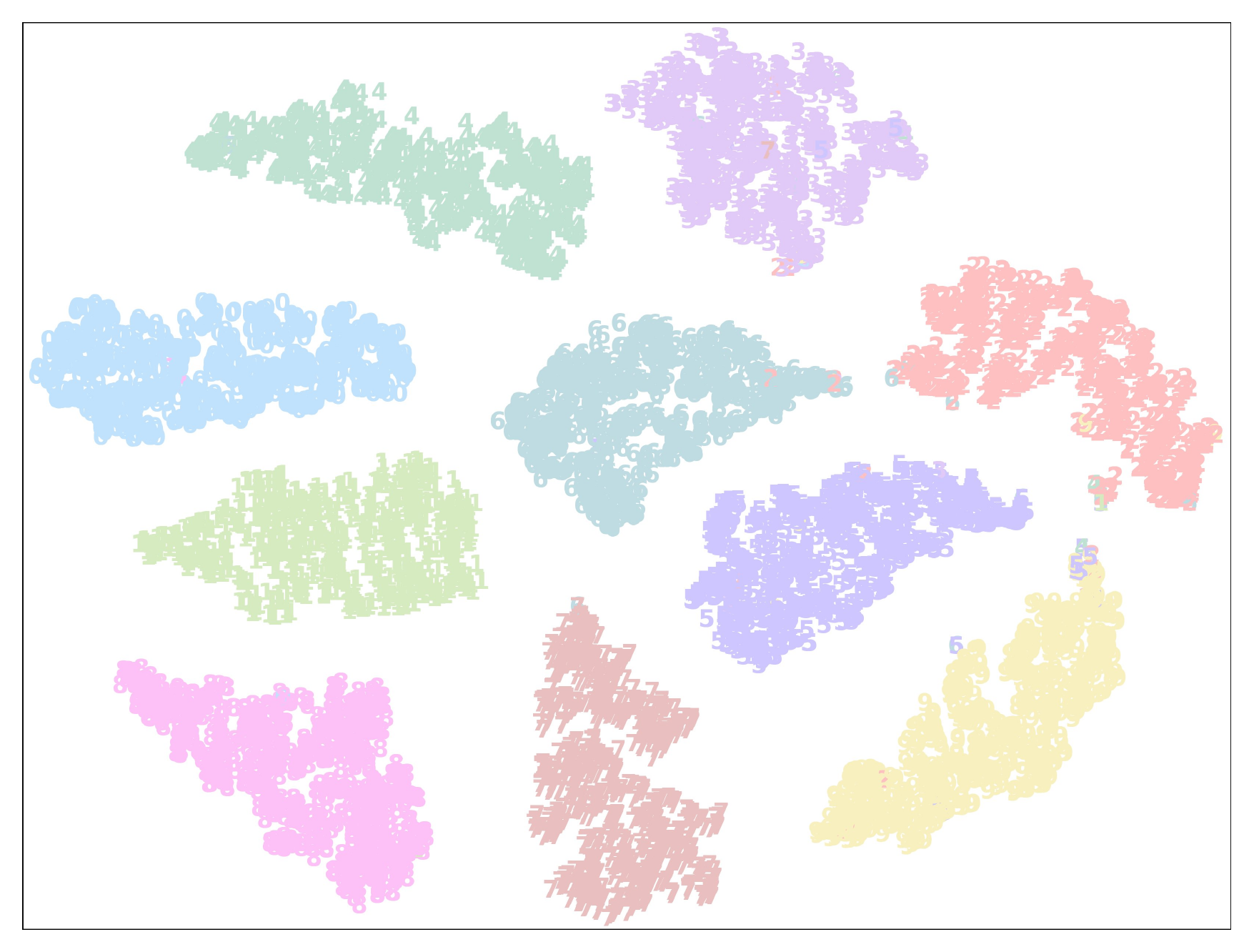}
        \caption{Epoch 40}
        \label{fig:Epoch401}
    \end{subfigure}
    \hspace{0.02mm}
    \begin{subfigure}[t]{0.145\textwidth}
        \centering
        \includegraphics[width=\linewidth]{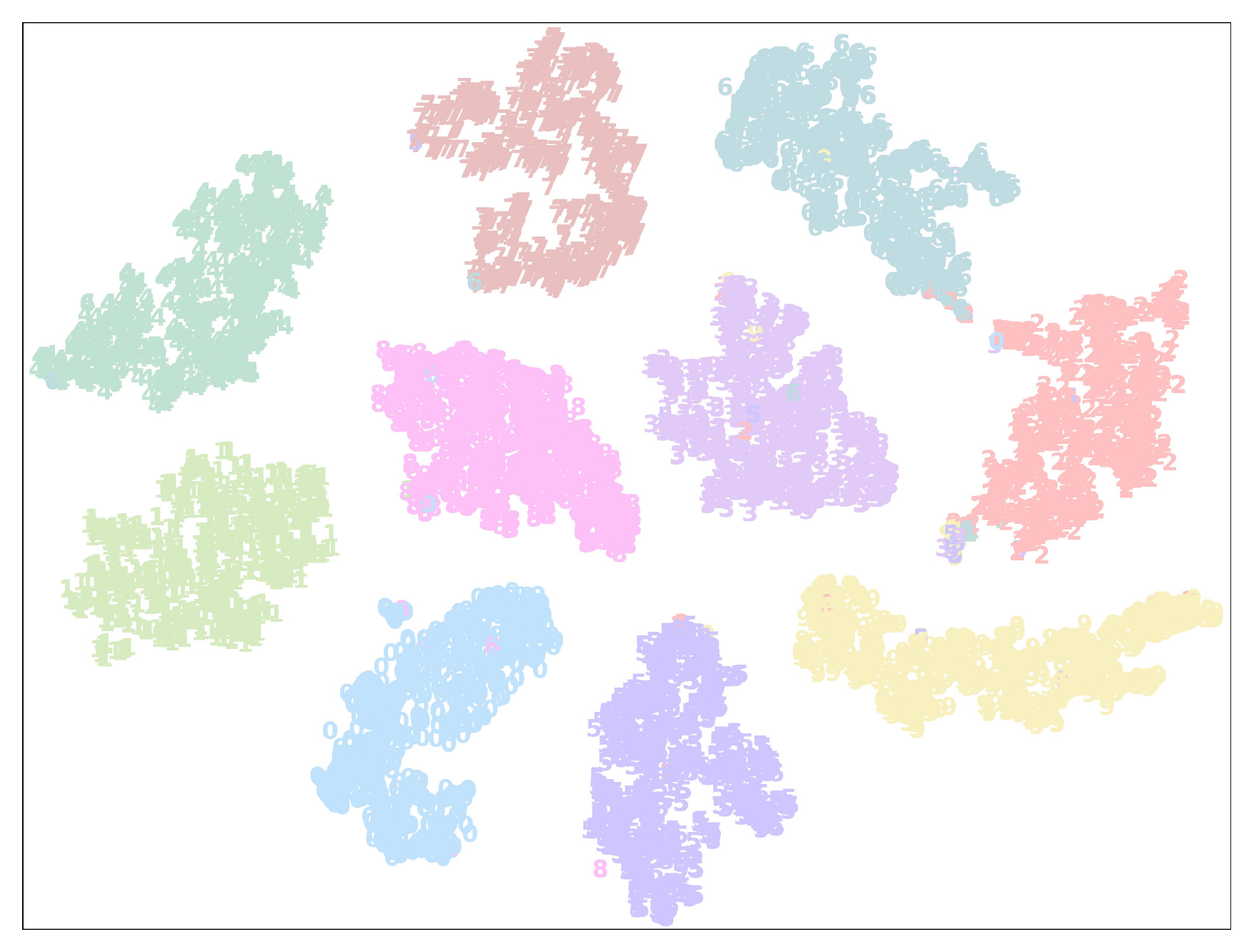}
        \caption{Epoch 80}
        \label{fig:Epoch801}
    \end{subfigure}
    
    \caption{2D visualization of the learned features at different epochs. (a–c): Amazon; (d–f): Fashion.}
    \label{fig:cluster_all}
\end{figure}

\textbf{Alignment mechanisms.} Table~\ref{tab:CL_SWD_accuracy_time} shows the effectiveness of $\mathcal{L}_{\mathrm{HHSW}}$ across Scene-15, Amazon, and YoutubeVideo. Compared to the hyperbolic instance-level contrastive loss $\mathcal{L}_{\text{HCL}}$ (see App.~\ref{appendix: Implementation Details}), our method consistently achieves superior or comparable results. (1) \textbf{Clustering performance:} $\mathcal{L}_{\text{HHSW}}$ outperforms $\mathcal{L}_{\text{HCL}}$ by 2.11\% (ACC) and 1.36\% (NMI) on Scene-15, and improves ACC by 6.95\% on YoutubeVideo, highlighting its advantage in modeling global multi-view semantics. (2) \textbf{Complexity:} 
$\mathcal{L}_{\text{HCL}}$ requires $\mathcal{O}(M^2B^2r)$ time due to pairwise comparisons, while $\mathcal{L}_{\mathrm{HHSW}}$ has a complexity of $\mathcal{O}(M^2 L B (r+\log B))$ due to efficient sorting over $L$ projections.
On large datasets YoutubeVideo, $\mathcal{L}_{\mathrm{HHSW}}$ achieves faster training than $\mathcal{L}_{\text{HCL}}$. Complexity details are provided in the App.~\ref{appendix: Complexity Analysis}.


\textbf{HHSW vs. GHSW.}
The key difference between HHSW and GHSW lies in their projection mappings~\cite{bonet2023hyperbolic}: HHSW uses a horospherical projection via the Busemann function, while GHSW employs a geodesic projection (see App.~\ref{appendix: OT-M} for details). We evaluate the impact of these two mapping strategies on the Fashion, Scene-15, and Amazon datasets using the Wasserstein distance with $p=2$. As shown in Table~\ref{tab:hhsw_ghsw_performance}, 
HHSW consistently outperforms GHSW across all evaluation metrics on all datasets. This demonstrates that the Busemann-based projection in HHSW better preserves semantic structures relevant to clustering.

\textbf{Temperature parameter.} In this part, we investigate the effect of the temperature parameter $\tau$ in Eq.~\ref{eq:label_contrastive_loss} on the clustering performance of WAH-MVC.
Figure~\ref{fig:temp_sensitivity_all} reports ACC and NMI on six different datasets as $\tau$ varies in the range $\{0.1, 1.0\}$. 
As shown, a moderate value range (\textit{e.g.}, $\tau=[0.3,0.5]$) generally yields better clustering performance, whereas excessively small or large values of $\tau$ deteriorate results. This is because $\tau$ balances sample discrimination hardness: too small a $\tau$ may over-emphasize hard negatives, while too large a $\tau$ will over-smooth the similarity distribution.  Therefore, choosing a proper $\tau$ improves both the stability and effectiveness of the contrastive learning process.

\textbf{Visualization.}
Figure~\ref{fig:loss_analize} plots the ACC, NMI, and total loss in Eq.~\ref{eq:total_loss} over training epochs on the Amazon and Scene-15 datasets. The total loss steadily decreases and quickly plateaus, while the ACC and NMI rise consistently before stabilizing, indicating that WAH-MVC converges reliably and maintains training stability.
Moreover, Figure~\ref{fig:cluster_all} visualizes the evolution of the learned features across epochs. This is realized using the t-SNE technique~\cite{van2008visualizing}, which projects high-dimensional features into a 2D space.  
As training progresses, intra-cluster compactness increases and inter-cluster ambiguity decreases, showing that WAH-MVC learns more discriminative multi-view representations despite semantic gaps.

\section{Conclusion}
In this paper, we propose WAH-MVC, a novel hyperbolic multi-view clustering framework that learns view-invariant representations by aligning cross-view semantic structures. To this end, we first design a view-specific hyperbolic encoder to capture the latent hierarchical features within each view. At its core, WAH-MVC performs SWD-based cluster-level alignment strategy on the Lorentz manifold, explicitly capturing the shared semantics among views and driving the model toward more discriminative and consistent cluster assignments. 
Extensive experiments and ablation studies on several benchmarking datasets demonstrate the superior performance of WAH-MVC and quantify the contribution of each component to the overall objective.

\section*{Acknowledgements}
This work was supported in part by the National Natural Science Foundation of China (62306127, 62020106012, 62332008), the Natural Science Foundation of Jiangsu Province (BK20231040), the Fundamental Research Funds for the Central Universities (JUSRP124015), the Key Project of Wuxi Municipal Health Commission (Z202318), the EU Horizon project ELIAS (101120237), the FIS project GUIDANCE (FIS2023-03251), and the National Key R\&D Program of China (2023YFF1105102, 2023YFF1105105).

\bibliography{references}
\clearpage
\AppendixIndependentNumbering

\AppendixSection{Optimization Process of WAH-MVC}
\label{appendix: Optimization Process of WAH-MVC}
Algorithm 1 presents the detailed optimization steps of the proposed WAH-MVC.

\begin{algorithm}[htbp]
\caption{Optimization Procedure for WAH-MVC}
\label{alg: Optimization Procedure for WAH-MVC}
\textbf{Input}: Multi-view dataset set $\mathcal{X}=\{X_m \in \mathbb{R}^{N \times D_m}\}_{m=1}^M$; training epochs $T$; loss weights $\alpha$, $\beta$, $\gamma$; temperature $\tau$; Wasserstein order $p$\\
\textbf{Output}: Labels $\mathbf{Y} = [y_1, \ldots, y_N]$
\begin{algorithmic}[1]
\STATE Initialize parameters of encoders $f_{\mathrm{enc}}(X_m; \theta_{enc}^m)$ for $m = 1,\dots,M$
\FOR{$t = 1$ to $T$}
    \STATE Extract hyperbolic embeddings $\{\widetilde{\mathcal{Z}}_{\text{hyp}}^{(m)} \in \mathbb{L}_K^{N \times r}\}_{m=1}^M$
    \STATE Compute soft cluster matrxs $\{\mathcal{A}^{(m)} \in \mathbb{R}^{N \times K}\}_{m=1}^M$
    \STATE Calculate redistributed targets $\{\mathbf{Q}^{(m)} \in \mathbb{R}^{N \times K}\}_{m=1}^M$ via Eq. 11 
    \STATE Update model parameters by minimizing total loss $\mathcal{L}$
\ENDFOR
\STATE Compute semantic cluster labels using Eq. 17 
\STATE \textbf{return} $\mathbf{Y}$
\end{algorithmic}
\end{algorithm}

\AppendixSection{ Description of Notations}
To be clear, the notations and corresponding definitions used in this paper are summarized in Table~\ref{tab:notations}.
\begin{table*}
    \centering
    \begin{tabular}{c|l}
    \midrule 
     Notation & Definition\\
    \midrule 
    $\mathcal{X}$ & The set of all data samples. \\
    $X_m$ & The $m$-th view of the multi-view data. \\
    $\mathcal{Z}_{hyp}^{(m)}$ & The Lorentz embedding of the $m$-th view. \\
    $\tilde{\mathcal{Z}}_{hyp}^{(m)}$ & The dimension-reduced Lorentz embedding of the $m$-th view. \\
    $\tilde{z}_{i}^{m}$ & The $i$-th sample in $\tilde{\mathcal{Z}}_{hyp}^{(m)}$. \\
    $\mathcal{A}^{(m)}$ & The soft cluster assignment matrix of the $m$-th view. \\
    $a_{ij}^{(m)}$ & The cluster assignment probability of sample $i$ to cluster $j$ in view $m$. \\
    $\mathbf{Q}^{(m)}$ & The target distribution computed from $\mathcal{A}^{(m)}$. \\
    $q_{k}^{m}$ & The $k$-th column of $Q^{(m)}$, corresponding to cluster $k$. \\
    $\boldsymbol{\Theta}$ & The set of projection directions in Lorentz space. \\
    $\boldsymbol{\theta}_{\ell}$ & The $\ell$-th projection direction. \\
    $B_{\boldsymbol{\theta}}$ & The Busemann projection along direction $\boldsymbol{\theta}$. \\
    $B_{\boldsymbol{\theta}_{\ell}}^m$ & The 1D projected distribution of view $m$ along direction $\boldsymbol{\theta}_{\ell}$. \\
    $N$ & The number of data samples. \\
    $M$ & The number of views. \\
    $K$ & The number of clusters. \\
    $L$ & The number of projection directions. \\
    $\mathcal{V}$ & The set of all view pairs, with $|\mathcal{V}| = M(M{-}1)/2$. \\
    $D_m$ & The Euclidean dimensionality of the $m$-th view. \\
    $r$ & The dimensionality of the latent Lorentz space. \\
    $\exp_{\mathbf{x}_\mathrm{o}}(\cdot)$ & The exponential map at Lorentz origin $\mathbf{x}_\mathrm{o}$. \\
    $\mathbf{x}_\mathrm{o}$ & The origin point in Lorentz space. \\
    \bottomrule  
    \end{tabular}
    \caption{Notations used in WAH-MVC}
    \label{tab:notations}
\end{table*}

\AppendixSection{Riemannian Geometry}
\label{appendix: Riemannian Geometry}

\AppendixSubsection{Preliminaries on Riemannian Geometry}
In this section, we provide a brief overview of Riemannian geometry and Riemannian manifolds. We denote the Euclidean inner product and norm for any real vectors $x, y \in \mathbb{R}^n$ as $\langle x, y \rangle$ and $\|x\|$, respectively.

An $n$-dimensional manifold $\mathcal{M}$
is a real, smooth space that can be locally approximated by an $n$-dimensional Euclidean space $\mathbb{R}^n$ at each point $x \in \mathcal{M}$. As a real-world analogy, the Earth can be globally modeled as a hypersphere (a smooth manifold), while locally, its surface can be approximated by a flat 2-dimensional plane $\mathbb{R}^2$.

At each point $x \in \mathcal{M}$, the corresponding tangent space $T_x\mathcal{M}$ is defined as the vector space of all possible directions in which one can tangentially pass through $x$ on a smooth path. This space is isomorphic to $\mathbb{R}^n$. A Riemannian metric $g = \{g_x\}_{x \in \mathcal{M}}$ is a collection of inner products $g_x: T_x\mathcal{M} \times T_x\mathcal{M} \to \mathbb{R}$, one for each tangent space $T_x\mathcal{M}$. The matrix form of the metric tensor at point $x$ is denoted as $G(x)$, and the inner product is given by
\begin{equation}
    g_x(u, v) = u^\top G(x) v, \quad \forall u, v \in T_x\mathcal{M}.
\end{equation}
The norm of a vector $z \in T_x\mathcal{M}$ is then defined as $\|z\|_x = \sqrt{\langle z, z \rangle_x}$.

Using the metric tensor, local geometric quantities such as angles, curve lengths, surface areas, and volumes can be defined and integrated to derive global geometric structures of the manifold. A Riemannian manifold is thus defined as the pair $(\mathcal{M}, g)$, where $\mathcal{M}$ is a differentiable manifold equipped with a Riemannian metric $g$.

The concept of a geodesic generalizes the notion of a straight line in Euclidean space to the manifold setting. It is defined as the shortest path between two points $x, y \in \mathcal{M}$ traced by a constant-speed curve $\gamma: [a, b] \to \mathcal{M}$ with constant speed. The length of the curve $\gamma$ is given by~\cite{ganea2018hyperbolic}
\begin{equation}
    L(\gamma) = \int_a^b \left| \gamma'(t) \right|_{\gamma(t)}^{\frac{1}{2}} \, dt
\end{equation}
and the geodesic distance between $x$ and $y$ is defined as
\begin{equation}
    d_{\mathcal{M}}(x, y) = \inf_{\gamma} L(\gamma),
\end{equation}
where the infimum is taken over all smooth curves $\gamma$ connecting $x$ and $y$.

To move a tangent vector along a geodesic while preserving its norm and direction, the parallel transport $P_{x\to y}: T_x\mathcal{M} \to T_y\mathcal{M}$ defines a linear isometry between tangent spaces.

To map a tangent vector back onto the manifold, the exponential map $\exp_x: T_x\mathcal{M} \to \mathcal{M}$ projects a vector from the tangent space onto the manifold along a geodesic. Its inverse, the logarithmic map $\log_x: \mathcal{M}\to T_x\mathcal{M}$, maps a point on the manifold back to the tangent space

\AppendixSubsection{Lorentz Model}
\label{appendix: Lorentz}
Here, we describe essential geometrical operations in the Lorentz model.
\paragraph{Tangent Space.} The tangent space at point $x \in \mathbb{L}_K^n$ is defined as:
\begin{equation}
    T_x \mathbb{L}_K^n := \left\{ y \in \mathbb{R}^{n+1} \mid \langle y, x \rangle_L = 0 \right\}.
\end{equation}

\paragraph{Logarithmic Maps.} 
The logarithmic map is the inverse of exponential map, mapping $y \in \mathbb{L}_K^n$ to $T_x \mathbb{L}_K^n$:
\begin{equation}
    \log_x^K(y) = \frac{
\operatorname{arccosh}(\beta)
}{
\sqrt{\beta^2 - 1}
}
\left( y - \beta x \right),\quad \beta = K \langle x, y \rangle_{\mathcal{L}}.
\end{equation}

\paragraph{Parallel Transport.}Parallel transport moves a tangent vector $v \in T_x \mathbb{L}_K^n$ along the geodesic from $x$ to $y \in \mathbb{L}_K^n$. In the Lorentz model, the operation is given by:
\begin{equation}
    \text{PT}^K_{x \rightarrow y}(v) = v - \frac{ \langle \log^K_x(y), v \rangle_L }{d_Ld_\mathcal{L}(x, y)} \left( \log^K_x(y) + \log^K_y(x) \right),
\end{equation}
where $d_{\mathcal{L}}(x, y)$ denotes the geodesic distance between $x$ and $y$ in the Lorentz model.

\subsection{Lorentz Neural Layers}
\paragraph{LorentzFC.}
To generalize neural networks to hyperbolic space, \citet{chen2021fully} 
proposed a fully-connected layer on the Lorentz manifold that supports both Lorentz boosts and rotations, enabling expressive geometric transformations in hyperbolic space. However, this formulation requires internal normalization and maintains strict orthogonality constraints, which increases implementation complexity.

To address these challenges, \citet{bdeir2023fully} 
introduced a simplified Lorentz fully-connected (LorentzFC) layer that directly maps Euclidean activations into the Lorentz manifold while preserving curvature-aware representations. Specifically, given an input vector $x \in \mathbb{L}_K^n$, weight matrix \( W \in \mathbb{R}^{m\times n + 1} \), bias \( b \in \mathbb{R}^{n} \), and nonlinearity \( \psi(\cdot) \) (e.g., identity or ReLU), the LorentzFC layer is defined as:
\begin{equation}
    y = \mathrm{LorentzFC}(x) = \left[    \begin{array}{c}
        \sqrt{ \| \psi(Wx + b) \|_2^2 - \frac{1}{K} } \\
        \psi(Wx + b)
    \end{array}
    \right].
\end{equation}
where the first component is the time-like coordinate and the remaining components form the spatial part. This construction guarantees that the output \( y \in \mathbb{L}_{K}^{n+1} \) lies on the Lorentz manifold, since
\begin{equation}
\begin{aligned}
\langle y, y \rangle_{\mathbb{L}} 
&= -y_0^2 + \| y_{1:n} \|_2^2 \\
&= -\left( \| \psi(Wx + b) \|^2 - \frac{1}{K} \right) + \| \psi(Wx + b) \|^2 \\
&= \frac{1}{K}
\end{aligned}
\end{equation}

This simplified design avoids the need for internal normalization, allowing external modules such as batch normalization to be applied, and facilitates efficient hyperbolic representation learning in practice.

\paragraph{LorentzMLR.}
Following the theoretical foundation presented in \citet{bdeir2023fully}
, we briefly review the closed-form expression for the minimum hyperbolic distance from a point $x \in \mathbb{L}_{K}^n$ to a decision hyperplane in Lorentz space, which underpins the Lorentz multinomial logistic regression (LorentzMLR) model.

Given $a \in \mathbb{R}$ and $z \in \mathbb{R}^n$, define $\alpha = \sqrt{-K}a$, the decision hyperplane is defined as:
\begin{equation}
\widetilde{H}_{z,a} = \left\{ x \in \mathbb{L}_K^n \ \middle| \ 
\cosh(\alpha) \langle z, x_s \rangle 
- \sinh(\alpha) \|z\| x_t = 0 \right\},
\end{equation}
where $x = (x_t, x_s)$ denotes the time and spatial components of $x$, respectively.

The minimum hyperbolic distance from $x \in \mathbb{L}_K^n$ to the hyperplane $\widetilde{H}_{z,a}$ is given by:
{
\small
\begin{equation}
    d_{\mathcal{L}}(x, \widetilde{H}_{z,a}) = \frac{1}{\sqrt{-K}} \left| \sinh^{-1}\left(\sqrt{-K} 
    \frac{ A \langle z, x_s \rangle - B \|z\| x_t }
    { \sqrt{ \| Az\|^2 - \left( B \|z\| \right)^2 } }
    \right) \right|,
\end{equation}
}
where $A=\cosh(\sqrt{-K}a)$ and $B=\sinh(\sqrt{-K}a)$. Building on this, \citet{bdeir2023fully}
extended the classical MLR model to Lorent space. For an input $x \in \mathbb{L}_{K}^n$ and a classification task with $C$ classes, each class $c \in \{1, \ldots, C\}$ is associated with parameters $(a_c, z_c)$, where $a_c \in \mathbb{R}$ and $z_c \in \mathbb{R}^n$. The logit score for class $c$ is defined as:
{
\small
\begin{equation}
\begin{aligned}
    v_{z_c, a_c}(x) &= \frac{1}{\sqrt{-K}} \operatorname{sign}(\alpha_c) \cdot \beta_c \cdot \left| \sinh^{-1} \left( \sqrt{-K} \cdot \frac{\alpha_c}{\beta_c} \right) \right|, \\
    \alpha_c &= \cosh(\sqrt{-K}a_c) \langle z_c, x_s \rangle - \sinh(\sqrt{-K}a_c) \|z_c\| x_t, \\
    \beta_c &= \sqrt{ \| \cosh(\sqrt{-K}a_c) z_c \|^2 - \left( \sinh(\sqrt{-K}a_c) \|z_c\| \right)^2 }.
\end{aligned}
\end{equation}
}

The predicted class probabilities are computed by the softmax function:
\begin{equation}
    p(y = c \mid x) = \frac{\exp\left(v_{z_c, a_c}(x)\right)}{\sum_{c'=1}^{C} \exp\left(v_{z_{c'}, a_{c'}}(x)\right)}.
\end{equation}

This Lorentz formulation preserves the underlying hyperbolic geometry and enables expressive, geometrically consistent decision boundaries, corresponding to geodesically curved hyperplanes in Lorentz space.

\AppendixSection{Wasserstein distance on Manifolds}
\label{appendix: OT-M}
The Wasserstein distance quantifies distributional divergence via optimal mass transport but suffers from high computational complexity of $\mathcal{O}(n^3 \log n)$ due to the need to solve a global transport plan over an $n \times n$ cost matrix
, limiting its scalability. To address this, the sliced Wasserstein distance (SWD)
projects distributions onto multiple random one-dimensional subspaces, where 1D Wasserstein distances can be efficiently computed via sorting with $\mathcal{O}(n \log n)$ complexity, reducing the overall cost to $\mathcal{O}(L \cdot n \log n)$ for $L$ projections. This efficiency has led to widespread use of SWD in applications such as generative modeling
and gradient flows
. To further extend its applicability, recent works have generalized SWD to non-Euclidean domains: \citet{pmlr-v202-rustamov23a}
introduced a spectral variant on compact manifolds via Laplace-Beltrami eigendecomposition, while  \citet{bonet2023spherical}
proposed an intrinsic formulation for the sphere. In hyperbolic geometry, where negative curvature introduces distinct challenges,  \citet{bonet2023hyperbolic} 
developed hyperbolic SWD variants based on geodesic and horospherical projections, enabling geometry-aware pseudo-distances for effective distribution alignment in tasks such as sampling and classification.

\AppendixSubsection{Details of HHSW and GHSW Distance}
\label{appendix: HHSW}
In this part, we provide the implementation details of the HHSW and GHSW distances, both of which are defined on the Lorentz model of hyperbolic space $\mathbb{L}_K^n$


\paragraph{Busemann Function on HHSW.}Each feature point $\tilde{z}_i^{m} \in \tilde{\mathcal{Z}}_{hyp}^{(m)}$ is projected from the hyperbolic manifold $\mathbb{L}_K^n$ to the real line via the Busemann projection along a given direction $\theta_\ell$, defined as:

\begin{equation}
\mathcal{B}_{\theta_\ell}(\tilde{z}_i^{m}) = \log \left( - \left\langle \tilde{z}_i^m, \mathbf{x}_\mathrm{o} + \theta_\ell \right\rangle_{\mathbb{L}} \right),
\label{eq: Horospherica projection}
\end{equation}
where $\mathbf{x}_\mathrm{o}$ is the origin of the Lorentz space and $\theta_\ell \in T_{\mathbf{x}_\mathrm{o}} \mathbb{L}_K^n \cap \mathbb{S}^{n-1}$ is a unit vector in the tangent space orthogonal to $\mathbf{x}_\mathrm{o}$. The scalar output $\mathcal{B}_{\theta_\ell}(\tilde{z}_i^{m})$ quantifies the signed distance from $\tilde{z}_i^{m}$ to a family of horospheres orthogonal to direction $\theta_\ell$.

\paragraph{Geodesic Projection on GHSW.}In contrast to HHSW, GHSW employs a geodesic projection to flatten hyperbolic points to the real line. Given a feature point $\tilde{z}_i^m$, the geodesic projection along direction $\theta_\ell$ is defined as:
\begin{equation}
P_{\theta_\ell}(\tilde{z}_i^m) = \operatorname{arctanh} \left( \frac{-\langle \tilde{z}_i^m, \theta_\ell \rangle_{\mathbb{L}}}{\langle \tilde{z}_i^m, \mathbf{x}_\mathrm{o} \rangle_{\mathbb{L}}} \right).
\label{eq: geodesic projection}
\end{equation}

This projection preserves geodesic distances by quantifying the hyperbolic displacement of $z_i^m$ along the direction $\theta_\ell$ from the origin $\mathbf{x}_\mathrm{o}$.

\paragraph{GHSW Distance.}Similarly, the GHSW distance is defined by integrating the Wasserstein distances of projected distributions:

\begin{equation}
\text{GHSW}_{\theta}^p(\mu, \nu) = \int_{T_{\mathbf{x}_\mathrm{o}}\mathbb{L}^n \cap \mathbb{S}^{n-1}} \mathcal{W}_p(P_{\theta}\#\mu, P_{\theta}\#\nu) \, d\lambda(\theta),
\end{equation}
where $P_{\theta}$ denotes the geodesic projection, and $\mathcal{W}_p$ is calculated in the same manner as in Eq. 7.

\paragraph{Comparison.}The key distinction between HHSW and GHSW lies in their projection mappings: HHSW leverages Busemann functions to encode distances to horospheres, while GHSW uses geodesic coordinates based on arctanh distances. Although both methods preserve curvature-aware properties of the Lorentz manifold, HHSW often enjoys better numerical stability and computational efficiency in high dimensions due to the simpler logarithmic form of its projection function.

\AppendixSection{Experimental Details}
\label{appendix:Experimental}

\AppendixSubsection{Evaluation Metrics}
\label{appendix: Evaluation Metrics}

We evaluate clustering performance using two standard metrics: Accuracy (ACC) and Normalized Mutual Information (NMI).

\begin{itemize}
    \item \textbf{ACC} measures the proportion of correctly assigned samples after finding the optimal one-to-one mapping between predicted clusters and ground-truth classes:
    \begin{equation}
        \text{ACC} = \max_{\pi \in \mathcal{P}} \frac{1}{N} \sum_{i=1}^{N} \mathbb{I}\big( y_i = \pi(c_i) \big),
    \end{equation}
    where $y_i$ and $c_i$ denote the ground-truth and predicted cluster labels of sample $i$, respectively, and $\pi$ is the permutation mapping between clusters and true labels.

    \item \textbf{NMI} quantifies the mutual dependence between predicted cluster assignments and ground-truth labels. It is invariant to label permutations and evaluates both completeness and homogeneity:
    \begin{equation}
        \text{NMI} = \frac{2 \times I(Y; C)}{H(Y) + H(C)},
    \end{equation}
    where $I(Y; C)$ is the mutual information between the true labels $Y$ and predicted clusters $C$, and $H(\cdot)$ denotes the Shannon entropy.
\end{itemize}

Higher values indicate better clustering performance, jointly reflecting alignment and consistency between predicted clusters and ground-truth classes.

\begin{figure*}[t]
    \centering
    \includegraphics[width=\linewidth]{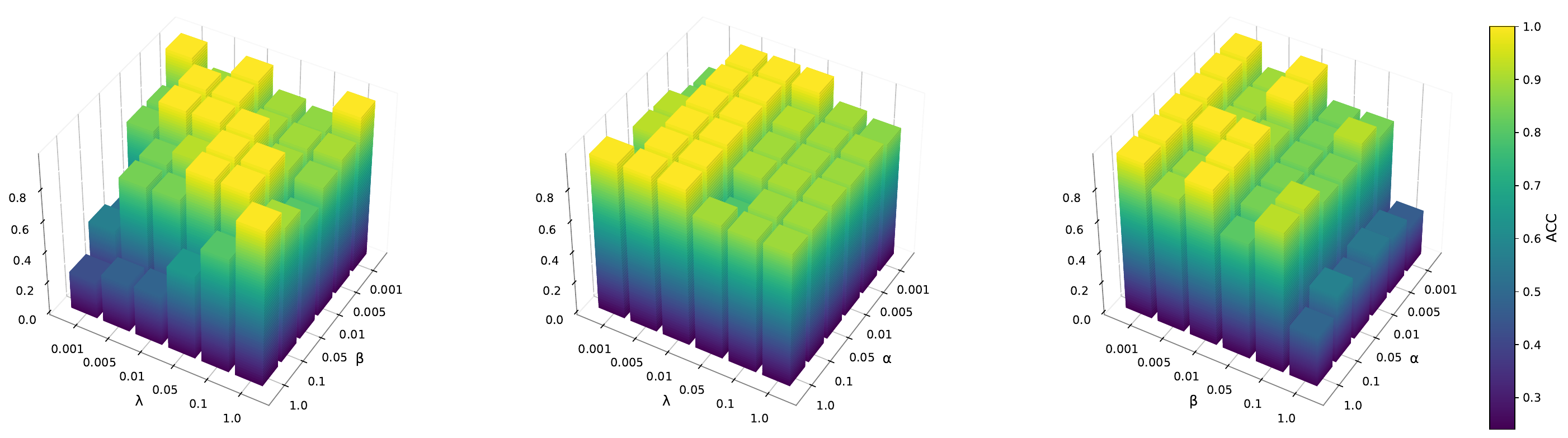}
   \caption{Accuracy comparison under different values of loss weight on the Amazon dataset.}
    \label{fig:parameter_sensitivity}
\end{figure*}
\AppendixSubsection{Implementation Details}
\label{appendix: Implementation Details}
For the vector-based input in the YouTubeVideo dataset, the encoder consists of a fully connected network with three Euclidean layers of size 256–512–1024, followed by an exponential map $\exp_{\mathbf{x}_\mathrm{o}}(\cdot)$ that projects the feature onto the Lorentz manifold $\mathbb{L}_K^{1025}$. The mapped feature is subsequently processed by two LorentzFC layers of size 1025–512–256 to obtain the final hyperbolic representation of dimension 256.

For the image inputs (MNIST-USPS, COIL-10, Amazon, Fashion, and Scene), we adopt an AlexNet-style encoder composed of two convolutional layers followed by adaptive average pooling. The resulting $64 \times 7 \times 7$ feature map is flattened and passed through an FC head. Specifically, for MNIST-USPS, COIL-10, Amazon, and Fashion, the FC head is used with the architecture $[64 \times 7 \times 7 \rightarrow 1024 \rightarrow 512 \rightarrow 256]$, followed by a Lorentz projection and two LorentzFC layers of sizes 257, 128, and 64. For the Scene dataset, a simplified FC head $[64 \times 7 \times 7 \rightarrow 1024 \rightarrow 1024]$ is employed, and the Lorentz representation is further reduced via two LorentzFC layers of size 1025–512–256. In all cases, the exponential map $\exp_{\mathbf{x}_\mathrm{o}}(\cdot)$ is applied to project the Euclidean feature into the Lorentz manifold before Lorentz-space dimensionality reduction. 

We set the Lorentz curvature parameter as $K = -1$ throughout all modules. For both HHSW and GHSW modules, we use the sliced Wasserstein power $p = 2$ and the projection direction dimension $L = 128$. Other hyperparameters, such as the temperature $\tau$ and the loss coefficients $\alpha$, $\beta$, and $\gamma$, are selected within appropriate ranges as shown in the experimental section. Full configuration details are available in the released code.

\AppendixSubsection{Complexity Analysis: $\mathcal{L}_{\text{HCL}}$ VS. $\mathcal{L}_{\text{HHSW}}$}
\label{appendix: Complexity Analysis}
We compare the time complexity of two multi-view alignment losses employed in hyperbolic representation learning: (1) Lorentz contrastive loss $\mathcal{L}_{\text{HCL}}$, and (2) HHSW alignment loss $\mathcal{L}_{\mathrm{HHSW}}$. The comparison is conducted under the assumption that each view contains $B$ samples in a batch, with Lorentz embedding dimensionality $r$, and the number of views is $M$.

\textbf{Lorentz Contrastive Loss:}
Given a pair of embeddings $\tilde{\mathbf{z}}_i^m, \tilde{\mathbf{z}}_i^n \in \mathbb{L}_K^r$ from two different views corresponding to the same sample, we define the Lorentz contrastive loss based on the pairwise Lorentz distance. Let $d_{\mathcal{L}} (\cdot, \cdot)$ denote the Lorentz distance in $\mathbb{L}_K^r$, and let $\tau$ be a temperature scaling factor. The loss over a batch of $B$ samples is formulated as:

\begin{equation}
\mathcal{L}_{\text{HCL}} = -\frac{1}{B} \sum_{i=1}^{B} \log \frac{
\exp\left(-\frac{1}{\tau}d_{\mathcal{L}}(\tilde{\mathbf{z}}_i^m, \tilde{\mathbf{z}}_i^n)\right)
}{
\sum_{\substack{j=1 \\ j \ne i}}^{B} \exp\left(-\frac{1}{\tau} d_{\mathcal{L}}(\tilde{\mathbf{z}}_i^m, \tilde{\mathbf{z}}_j^n) \right)
}
\end{equation}
For each pair of views $(m, n)$, the Lorentz contrastive loss computes a pairwise distance matrix between all samples from both views. Specifically, this involves computing $B\times B$ Lorentz distances, each requiring $\mathcal{O}(d)$ operations due to the Minkowski inner product and square-root operations. Given that there are $\mathcal{O}(M^2)$ such view pairs, the overall complexity is \textbf{$\mathcal{O}(M^2B^2d)$}. This quadratic dependence on the batch size $B$ can become a computational bottleneck when $B$ is large. 

\textbf{HHSW alignment loss:}
The HHSW loss estimates the distributional discrepancy between views using projections onto $L$ randomly sampled directions in the tangent space. For each projection, Busemann functions are computed for all $B$ samples in both views, each with a cost of $\mathcal{O}(d)$. After projection, the 1D Wasserstein distance is estimated by sorting the projected values, incurring a cost of $\mathcal{O}(B \log B)$ per direction. Hence, for $L$ directions, the per-view-pair complexity is \textbf{$\mathcal{O}(LB(d+\log (B))$}. Aggregating over all $\binom{M}{2}$ view pairs results in an overall complexity of \textbf{$\mathcal{O}(M^2LB(d+\log (B))$}. This complexity scales linearly with batch size $B$ and is typically lower than the Lorentz contrastive loss when $B$ is moderate to large and the number of projections $L$ is reasonably small (e.g., $L \leq 128$).

\begin{table}[t]
\centering
\setlength{\tabcolsep}{1mm}
{\fontsize{9}{10}\selectfont
\begin{tabular}{c|cccccccc}
\toprule
$p$ & 1  & 2  & 3 & 5 & 10 & 30 & 50 & 100\\
\midrule
ACC  & 72.00  & 74.38  & 71.13 & 72.31 & 71.68 & 71.36 & 70.95 & 70.84 \\
NMI & 73.71  & 74.68 & 73.89 & 73.15 & 73.41 & 73.60 & 72.61 & 72.56\\
\bottomrule
\end{tabular}
}
\caption{Ablation of HHSW order $p$ on Scene-15}
\label{tab:ablation_p_hhsw}
\end{table}

\AppendixSubsection{Supplementary experiments}
\label{appendix: Supplementary experiments}
\textbf{Wasserstein distance parameter $p$.}
We study the impact of the Wasserstein order $p$ on the Scene-15 dataset. As shown in Table~\ref{tab:ablation_p_hhsw}, performance improves with $p$ up to 2, then gradually degrades as $p$ increases further. This indicates that a moderate $p$ (e.g., $p{=}2$)  achieves better clustering results, while overly large values hurt performance. This is likely because higher $p$ magnifies sensitivity to outliers and long-range discrepancies, amplifying noise and reducing stability

\textbf{Trade-off Parameter Sensitivity.} To further assess the influence of the trade-off parameters in Eq. 16, we perform parameter sensitivity experiments on the Amazon dataset by fixing one parameter among $\alpha$, $\beta$, and $\gamma$, while varying the other two within the range ${0.001, 0.005, 0.01, 0.05, 0.1, 1.0}$. As shown in Fig.~\ref{fig:parameter_sensitivity}, when $\alpha$ is fixed, the performance is highly sensitive to the choice of $\beta$ and $\gamma$, with moderate values (e.g., 0.01–0.05) generally achieving superior clustering accuracy. Fixing $\beta$ reveals that appropriate tuning of $\alpha$ and $\gamma$ remains crucial, where low $\alpha$ and mid-range $\gamma$ yield favorable results. Similarly, when $\gamma$ is fixed, clustering performance improves with increasing $\beta$ up to a certain point before declining, while $\alpha$ exhibits a more stable influence. These results indicate that $\beta$ and $\gamma$ have a more pronounced impact on performance compared to $\alpha$. That careful tuning of these parameters is essential to exploit the benefits of the proposed loss formulation fully.
\begin{table}
    \centering
    \begin{tabular}{c|ccccc}
    \midrule 
    Missing Rate & 0 & 0.1 & 0.3 & 0.5 & 0.7\\
    \midrule 
    ACC  & 29.40 & 23.29 & 19.49 & 16.67 & 15.93 \\
    NMI  & 22.74 & 21.86 & 16.22 & 12.41 & 8.29 \\
    \bottomrule  
    \end{tabular}
    \caption{The clustering performance of incomplete views on the YoutubeVideo dataset}
    \label{tab:incom}
\end{table}

\textbf{Experiments on Incomplete Views.}
To evaluate the robustness of our method in incomplete multi-view scenarios, we simulated varying degrees of view loss on the YouTube Video dataset (missing rates of 0.1, 0.3, 0.5, and 0.7). As shown in Table~\ref{tab:incom}, clustering performance systematically degrades with increasing missing rates. However, these experiments also reveal limitations of our model: First, even with complete views (missing rate 0), the model's initial performance (ACC 29.40\%, NMI 22.74\%) still leaves room for improvement, reflecting the inherent challenges of this dataset and its task. Second, performance is highly sensitive to view loss, with particularly high missing rates (0.5 and 0.7) experiencing a sharp decline (e.g., ACC drops to 16.67\% and 15.93\%, and NMI drops to 12.41\% and 8.29\%), indicating the model's difficulty in leveraging extremely sparse view information. 

However, the core value of this experiment lies in: 
(1) It systematically quantifies the impact of missing views on performance, providing an empirical basis for understanding the failure boundary of the model; (2) The results show that at low and medium missing rates (such as 0.1 and 0.3), the model can still maintain a certain performance using the remaining views (such as ACC 19.49 and NMI 16.22 at a missing rate of 0.3), demonstrating a certain information utilization capability and robustness; (3) This experimental design that simulates real-world defects (such as sensor failures) under controllable conditions is crucial for promoting the application of multi-view learning in real-world scenarios, and points out the direction for future improvements (such as enhancing the missing view processing mechanism).



\end{document}